\documentclass[10pt,journal,compsoc]{IEEEtran}

\usepackage{graphicx}
\graphicspath{{Figures/}}
\usepackage{cite}
\usepackage[tight,footnotesize]{subfigure}
\usepackage[T1]{fontenc}

\usepackage[table,xcdraw]{xcolor}
\usepackage{color}
\definecolor{darkblue}{rgb}{0.0,0.0,0.0}
\usepackage{hyperref}
\hypersetup{colorlinks=true,linkcolor=blue, urlcolor=darkblue,anchorcolor=darkblue, citecolor=blue}
\usepackage{booktabs}
 \usepackage{multirow}
\usepackage{amsmath}
\usepackage{amssymb}
\usepackage{xspace}

\newcommand{\ie}{i.e.,\xspace}
\newcommand{\eg}{e.g.,\xspace}
\newcommand{\etal}{et al.\xspace}
\newcommand{\paratitle}[1]{\vspace{0.8ex}\noindent \textbf{#1}}
\newcommand{\tabincell}[2]{\begin{tabular}{@{}#1@{}}#2\end{tabular}}

\begin{document}
\title{A Survey on Deep Learning for \\ Named Entity Recognition}

\author{Jing Li,
	Aixin Sun,
	Jianglei Han,
	and Chenliang Li
	\IEEEcompsocitemizethanks{\IEEEcompsocthanksitem J. Li is with the Inception Institute of Artificial Intelligence, United Arab Emirates. This work was done when the author was with Nanyang Technological University, Singapore. E-mail: jli030@e.ntu.edu.sg. 
	\IEEEcompsocthanksitem A. Sun is with School of Computer Science and Engineering, Nanyang Technological University, Singapore. E-mail: axsun@ntu.edu.sg.
	\IEEEcompsocthanksitem J. Han is with SAP, Singapore. E-mail: ray.han@sap.com.
	\IEEEcompsocthanksitem C. Li is with School of Cyber Science and Engineering, Wuhan University, China. E-mail:cllee@whu.edu.cn.}
	\thanks{Accepted in IEEE TKDE.}}

\markboth{IEEE TRANSACTIONS ON KNOWLEDGE AND DATA ENGINEERING, 2020}
{Li \etal: A Survey on Deep Learning for Named Entity Recognition}

\IEEEtitleabstractindextext{%

\begin{abstract}
Named entity recognition (NER) is the task to identify  mentions of rigid designators from text belonging to predefined semantic types such as person, location, organization etc. 
NER always serves as the foundation for many natural language applications such as question answering, text summarization, and machine translation.
Early NER systems got a huge success in achieving good performance with the cost of human engineering in designing domain-specific features and rules. 
In recent years, deep learning, empowered by continuous real-valued vector representations and semantic composition through nonlinear processing, has been employed in NER systems, yielding stat-of-the-art performance.
In this paper, we provide a comprehensive review on existing deep learning techniques for NER.
We first introduce NER resources, including tagged NER corpora and off-the-shelf NER tools.
Then, we systematically categorize existing works based on a taxonomy along three axes: distributed representations for input, context encoder, and tag decoder.
Next, we survey the most representative methods for recent applied techniques of deep learning in new NER problem settings and applications.
Finally, we present readers with the challenges faced by NER systems and outline future directions in this area.
\end{abstract}

}

\maketitle

\IEEEdisplaynontitleabstractindextext

\IEEEpeerreviewmaketitle

\section{Introduction}\label{sec:introduction}

\IEEEPARstart{N}{amed} Entity Recognition (NER) aims to recognize mentions of rigid designators from text belonging to predefined semantic types such as person, location, organization etc~\cite{nadeau2007survey}.
NER not only acts as a standalone tool for information extraction (IE), but also plays an essential role in a variety of natural language processing (NLP) applications such as text understanding \cite{DBLP:conf/acl/ZhangHLJSL19,cheng2019attending}, information retrieval~\cite{guo2009named,petkova2007proximity}, automatic text summarization~\cite{larsen1999trainable},  question answering~\cite{molla2006named}, machine translation~\cite{babych2003improving}, and knowledge base construction~\cite{etzioni2005unsupervised} etc.

\paratitle{Evolution of NER.}
The term ``\textit{Named Entity}'' (NE) was first used at the sixth Message Understanding Conference (MUC-6)~\cite{grishman1996message}, as the task of identifying names of organizations, people and geographic locations in text, as well as currency, time and percentage expressions.
Since MUC-6 there has been increasing interest in NER, and various scientific events (\eg CoNLL03~\cite{tjong2003introduction}, ACE~\cite{doddington2004automatic}, IREX~\cite{demartini2009overview}, and TREC Entity Track~\cite{balog2010overview}) devote much effort to this topic.

Regarding the problem definition, Petasis \etal~\cite{petasis2000automatic} restricted the definition of named entities: ``A NE is a proper noun, serving as a name for something or someone''. This restriction is justified by the significant  percentage of proper nouns present in a corpus.
 Nadeau and Sekine~\cite{nadeau2007survey} claimed that  the word ``\textit{Named}'' restricted the task to only those entities for which one or many \textit{rigid designators} stands for the referent. Rigid designator, defined in~\cite{kripke1972naming}, include proper
names and  natural kind terms like biological species and substances.
 Despite the various definitions of NEs, researchers  have reached common consensus on the types of NEs to recognize.
 We generally divide NEs into two categories: generic NEs (\eg person and location) and domain-specific NEs (\eg proteins, enzymes, and genes).
In this paper, we mainly focus on generic NEs in English language. We do not claim this article to be exhaustive or representative of all NER works on all languages.

As to the techniques applied in NER, there are four main streams:
1)  Rule-based approaches, which do not need annotated data as they rely on hand-crafted rules;
2) Unsupervised learning approaches, which rely on unsupervised algorithms without hand-labeled training examples;
3) Feature-based supervised learning approaches, which rely on  supervised learning algorithms with careful feature engineering;
4) Deep-learning based approaches, which automatically discover representations needed for the classification and/or detection from raw input in an end-to-end manner.  We brief 1), 2) and 3), and review 4) in detail.

\paratitle{Motivations for conducting this survey.}
In recent years, deep learning (DL, also named deep neural network) has attracted significant attention due to its success in various  domains.
Starting with Collobert \etal~\cite{collobert2011natural},
DL-based NER systems with minimal feature engineering have been flourishing.
Over the past few years, a considerable number of studies have applied deep learning to NER and successively advanced the state-of-the-art performance~\cite{collobert2011natural,huang2015bidirectional,lample2016neural,chiu2016named,peters2017semi}.
This trend motivates us to conduct a survey to report the current status of deep learning techniques in NER research.
By comparing the choices of DL architectures, we aim to identify factors affecting NER performance as well as issues and challenges.

On the other hand, although NER studies have been thriving for a few decades, to the best of our knowledge, there are few reviews in this field so far.
Arguably the most established one was published by Nadeau and Sekine~\cite{nadeau2007survey} in 2007.
This survey presents an overview of the technique trend  from hand-crafted rules towards machine learning.
Marrero \etal~\cite{marrero2013named} summarized NER works from the perspectives of fallacies, challenges and opportunities in 2013.
Then Patawar and Potey~\cite{patawar2015approaches}  provided a short review in 2015.
The two recent short surveys are  on new domains~\cite{saju2017survey} and complex entity mentions~\cite{dai2018recognizing}, respectively.
In summary, existing surveys mainly cover feature-based machine learning models, but not the modern DL-based NER systems.
More germane to this work are the two recent surveys~\cite{yadav2018survey,goyal2018recent} in 2018.
Goyal \etal~\cite{goyal2018recent} surveyed developments and progresses made in NER.
However, they did not include recent advances of deep learning techniques.
Yadav and Bethard~\cite{yadav2018survey} presented a short survey of recent advances in NER based on representations of  words in sentence.
This survey focuses more on the distributed representations for input (\eg char- and word-level embeddings) and do not review the context encoders and tag decoders. The recent trend of applied deep learning on NER tasks (\eg multi-task learning, transfer learning, reinforcement leanring and adversarial learning) are not in their servery as well.

\paratitle{Contributions of this survey.}
We intensely review applications of deep learning techniques  in NER, to enlighten and guide researchers and practitioners  in this area. Specifically, we consolidate NER corpora, off-the-shelf NER systems (from both academia and industry) in a tabular form, to provide useful resources for NER research community.
We then present a comprehensive survey on deep learning techniques for NER. To this end, we propose a new taxonomy, which systematically organizes DL-based NER approaches along three axes: distributed representations for input, context encoder (for capturing contextual dependencies for tag decoder), and tag decoder (for predicting labels of words in the given sequence). In addition, we also survey the most representative methods for recent applied deep learning techniques in new NER problem settings and applications. Finally, we present readers with the challenges faced by NER systems and outline future directions in this area.

\section{Background}
\label{sec:background}
We first give a formal formulation of the NER problem.
We then introduce the widely-used NER datasets and tools. Next, we detail the evaluation metrics and summarize the traditional approaches to NER.

\subsection{What is NER?}

A named entity is a word or a phrase that clearly identifies one item from a set of other items that have similar attributes \cite{sharnagat2014named}. Examples of named entities are organization, person, and location names in general domain; gene, protein, drug and disease names in biomedical domain. NER is the process of locating and classifying named entities in text into predefined entity categories.

\begin{figure}[t]
	\centering	
	\includegraphics[width=0.8\columnwidth,draft=false]{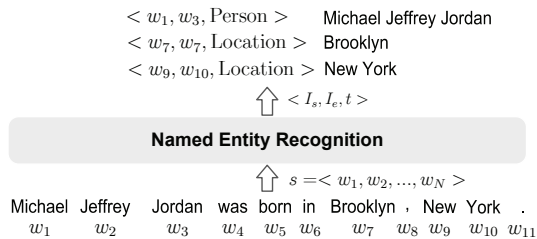}	
	\vspace{-2mm}
	\caption{An illustration of the named entity recognition task.}
	\label{fig:nerexample}
\end{figure}

Formally, given a sequence of tokens $s= \langle w_1, w_2,..., w_N\rangle$, NER is to output a list of tuples $\langle I_s, I_e, t \rangle$, each of which is a named entity mentioned in $s$. Here, $I_s \in [1, N]$ and $I_e \in [1, N] $ are the start and the end indexes of a named entity mention; $t$ is the entity type from a predefined category set. Figure~\ref{fig:nerexample} shows an example where a NER system recognizes three named entities from the given sentence. 
When NER was first defined in MUC-6~\cite{grishman1996message}, the task is to recognize names of people, organizations, locations, and time, currency, percentage expressions in text. Note that the task focuses on a small set of coarse entity types and one type per named entity. We call this kind of NER tasks as coarse-grained NER~\cite{grishman1996message,tjong2003introduction}. Recently, some fine-grained NER tasks~\cite{ling2012fine,ren2016afet,abhishek2017fine,lal2017sane,corro2015finet} focus on a much larger set of entity types where a mention may be assigned multiple fine-grained types.

NER acts as an important pre-processing step for a variety of downstream applications such as information retrieval, question answering, machine translation, etc.
Here, we use semantic search as an example  to illustrate the importance of NER in supporting various applications.
Semantic search refers to a collection of techniques, which enable search engines to understand the concepts, meaning, and intent behind the queries from users~\cite{balog2018entity}.
According to~\cite{guo2009named}, about 71\% of search queries contain at least one named entity.
Recognizing named entities in search queries would help us to better understand user intents, hence to provide better search results.
To incorporate named entities in search, entity-based language models~\cite{balog2018entity}, which consider individual terms as well as term sequences that have been annotated as entities (both in documents and in queries), have been proposed by Raviv \etal~\cite{raviv2016document}. There are also studies utilizing named entities for an enhanced user experience, such as query recommendation~\cite{boldi2008query}, query auto-completion~\cite{cai2016survey,bar2011context} and entity cards~\cite{saldanha2016entity,hasibi2017dynamic}.

\begin{table*}[t]
	\centering
	\caption{List of annotated datasets for English NER. ``\#Tags'' refers to the number of entity types.}	
	\label{tab:datasets}
	\begin{tabular}{ccccp{8.2cm}}
		\toprule
		Corpus & Year & Text Source & \#Tags & URL \\ \midrule
		MUC-6& 1995     &  Wall Street Journal           & 7       & \url{https://catalog.ldc.upenn.edu/LDC2003T13}    \\
		MUC-6 Plus&  1995    & Additional news to MUC-6            & 7       &   \url{https://catalog.ldc.upenn.edu/LDC96T10}  \\
		MUC-7&   1997  &     New York Times news         &   7     &    \url{https://catalog.ldc.upenn.edu/LDC2001T02}  \\
		CoNLL03&  2003    &  Reuters  news           & 4       & \url{https://www.clips.uantwerpen.be/conll2003/ner/}    \\
			ACE&  2000 - 2008    &  Transcripts,  news            & 7       & \url{https://www.ldc.upenn.edu/collaborations/past-projects/ace} \\
		OntoNotes&  2007 - 2012    &  Magazine, news, web, etc.            & 18       & \url{https://catalog.ldc.upenn.edu/LDC2013T19}    \\
		W-NUT&  2015 - 2018    &  User-generated text          & 6/10       & \url{http://noisy-text.github.io}    \\
		BBN&  2005    &  Wall Street Journal           & 64    & \url{https://catalog.ldc.upenn.edu/LDC2005T33}    \\
				WikiGold&  2009    &  Wikipedia        & 4    &  https://figshare.com/articles/Learning\_multilingual\_named\_entity \_recognition\_from\_Wikipedia/5462500  \\
			WiNER&  2012    &  Wikipedia        & 4    & \url{http://rali.iro.umontreal.ca/rali/en/winer-wikipedia-for-ner}    \\
			WikiFiger&  2012    &  Wikipedia        & 112    & \url{https://github.com/xiaoling/figer}    \\
			HYENA&  2012    &  Wikipedia        & 505    & https://www.mpi-inf.mpg.de/departments/databases-and-information-systems/research/yago-naga/hyena   \\

			N$^3$&  2014    &   News       & 3    & \url{http://aksw.org/Projects/N3NERNEDNIF.html}    \\	
				Gillick&  2016    &   Magazine, news, web, etc.        & 89    &\url{https://arxiv.org/e-print/1412.1820v2}    \\			
				FG-NER&  2018    &   Various        & 200    &\url{https://fgner.alt.ai/}    \\	
				NNE&  2019    &   Newswire        & 114    &\url{https://github.com/nickyringland/nested_named_entities}    \\	
				GENIA& 2004     &  Biology and clinical text        & 36   & \url{http://www.geniaproject.org/home}    \\
					GENETAG& 2005     &  MEDLINE         & 2   & \url{https://sourceforge.net/projects/bioc/files/}    \\
		FSU-PRGE& 2010     & PubMed and MEDLINE        & 5   & \url{https://julielab.de/Resources/FSU_PRGE.html}    \\
			NCBI-Disease & 2014     & PubMed        & 1   & \url{https://www.ncbi.nlm.nih.gov/CBBresearch/Dogan/DISEASE/}    \\
			BC5CDR & 2015     & PubMed        & 3   & \url{http://bioc.sourceforge.net/}    \\
			DFKI & 2018     & Business news and social media         & 7   & \url{https://dfki-lt-re-group.bitbucket.io/product-corpus/}    \\
			 \bottomrule
	\end{tabular}
\end{table*}

\subsection{NER Resources: Datasets and Tools}
High quality annotations are critical for both model learning and evaluation. In the following,  we summarize widely-used datasets and off-the-shelf tools for English NER.

A tagged corpus is a collection of documents that contain annotations of one or more entity types. Table~\ref{tab:datasets} lists some widely-used datasets with their data sources and number of entity types (also known as tag types).  Summarized in Table~\ref{tab:datasets}, before 2005, datasets were mainly developed by annotating news articles with a small number of entity types, suitable for coarse-grained NER tasks. After that, more datasets were developed on various kinds of text sources including Wikipedia articles, conversation, and user-generated text (\eg tweets and YouTube comments and StackExchange posts in W-NUT).  The number of tag types becomes significantly larger, \eg 505 in HYENA. We also list a number of domain specific datasets, particularly developed on PubMed and MEDLINE texts. The number of entity types ranges from 1 in NCBI-Disease to 36 in GENIA.

We note that many recent NER works report their performance on  CoNLL03 and OntoNotes datasets (see Table~\ref{tab:tabsurvey}). CoNLL03 contains annotations for Reuters news in two languages: English and German. The English dataset has a large portion of sports news with annotations in four entity types (Person, Location, Organization, and Miscellaneous)~\cite{tjong2003introduction}. The goal of the OntoNotes project was to annotate a large corpus, comprising of various genres (weblogs, news, talk shows, broadcast, usenet newsgroups, and conversational telephone speech)  with structural information (syntax and predicate argument structure) and shallow semantics (word sense linked to an ontology and coreference). 
There are 5 versions, from Release 1.0 to Release 5.0. The texts are annotated with 18 entity types.
We also note two Github repositores\footnote{\url{https://github.com/juand-r/entity-recognition-datasets} and \url{https://github.com/cambridgeltl/MTL-Bioinformatics-2016/tree/master/data}} which host some NER corpora.

\begin{table}[t]
\small
	\centering
		\caption{Off-the-shelf NER tools offered by academia and industry projects.}	
	\label{tab:nertools}
		\scalebox{0.9}{
	\begin{tabular}{lp{6.3cm}}
		\toprule
		NER System & URL \\ \midrule
		StanfordCoreNLP&   \url{https://stanfordnlp.github.io/CoreNLP/} \\
		OSU Twitter NLP&    \url{https://github.com/aritter/twitter_nlp} \\
		Illinois NLP&    \url{http://cogcomp.org/page/software/} \\
		NeuroNER& \url{http://neuroner.com/}    \\
		NERsuite&  \url{http://nersuite.nlplab.org/}   \\
		Polyglot& \url{https://polyglot.readthedocs.io}    \\
		Gimli& \url{http://bioinformatics.ua.pt/gimli}    \\
		\hline		
		spaCy& \url{https://spacy.io/api/entityrecognizer}    \\
		NLTK&  \url{https://www.nltk.org}   \\		
		OpenNLP&  \url{https://opennlp.apache.org/}   \\
		LingPipe&  \url{http://alias-i.com/lingpipe-3.9.3/}   \\		
		AllenNLP& \url{https://demo.allennlp.org/}    \\
		IBM Watson& https://natural-language-understanding-demo.ng.bluemix.net    \\		
		FG-NER& \url{https://fgner.alt.ai/extractor/}    \\
		Intellexer& \url{http://demo.intellexer.com/}    \\
		Repustate& https://repustate.com/named-entity-recognition-api-demo    \\
		AYLIEN& \url{https://developer.aylien.com/text-api-demo}    \\		
		Dandelion API& https://dandelion.eu/semantic-text/entity-extraction-demo   \\		
		displaCy& \url{https://explosion.ai/demos/displacy-ent}    \\		
		ParallelDots& https://www.paralleldots.com/named-entity-recognition   \\		
		TextRazor& https://www.textrazor.com/named \_entity\_recognition   \\		
		
 \bottomrule
	\end{tabular}}
\end{table}

There are many NER tools available online with pre-trained models. Table~\ref{tab:nertools} summarizes popular ones for English NER by academia (top) and industry (bottom).

\subsection{NER Evaluation Metrics}

NER systems are usually evaluated by comparing their outputs against human annotations. The comparison can be quantified by either exact-match or relaxed match.

\subsubsection{Exact-match Evaluation}

NER essentially involves two subtasks: boundary detection and type identification. 
In ``exact-match evaluation''~\cite{tjong2003introduction, pradhan2012conll,DBLP:journals/corr/abs-1904-10503}, a correctly recognized instance requires a system to correctly identify its boundary and type, simultaneously. 
More specifically,  the numbers of False positives (FP),  False negatives (FN) and True positives (TP) are used to compute Precision, Recall, and F-score.

\begin{itemize}
\item False Positive (FP): entity that is returned by a NER system but does not appear in the ground truth. 
\item False Negative (FN):  entity that is not returned by a NER system but appears in the ground truth. 
\item True Positive (TP): entity that is returned by a NER system and also appears in the ground truth. 
\end{itemize}

Precision refers to the percentage of your system results which are correctly recognized. 
Recall refers to the percentage of total entities correctly recognized by your system. 

\begin{equation}
\text{Precision} = \frac{\#  TP}{\#  (TP+FP)}  \quad\quad   \text{Recall} =  \frac{\#  TP}{\#  (TP+FN)} \notag
\end{equation}

A measure that combines precision and recall is the harmonic mean of precision and recall, the traditional F-measure or balanced F-score:
\begin{equation}
\text{F-score} = 2 \times \frac{\text{Precision}\times \text{Recall}}{\text{Precision}+ \text{Recall}} \notag
\end{equation}

In addition, the macro-averaged F-score and micro-averaged F-score both consider the performance across multiple entity types.  
Macro-averaged F-score independently calculates the F-score on different entity types, then takes the average of the F-scores. 
Micro-averaged F-score sums up the individual  false negatives, false positives and  true positives across all entity types then applies them to get the statistics.
The latter can be heavily affected by the quality of recognizing entities in large classes in the corpus.

\subsubsection{Relaxed-match Evaluation}

MUC-6~\cite{grishman1996message} defines a relaxed-match evaluation: a correct type is credited if an entity is assigned its correct type regardless its boundaries as long as there is an overlap with ground truth boundaries; a correct boundary is credited regardless an entity's type assignment. Then ACE~\cite{doddington2004automatic} proposes a more complex evaluation procedure.
It resolves a few issues like partial match and wrong type, and considers subtypes of named entities.
However, it is problematic because the final scores are comparable only when parameters are fixed~\cite{nadeau2007survey,patawar2015approaches,marrero2013named}.
Complex evaluation methods are not intuitive and make error analysis difficult.
Thus, complex evaluation methods are not widely used in recent studies.

\subsection{Traditional Approaches to NER}

Traditional approaches to NER are broadly classified into three main streams: rule-based, unsupervised learning, and feature-based supervised learning approaches~\cite{nadeau2007survey,yadav2018survey}.

\subsubsection{Rule-based Approaches}

Rule-based NER systems rely on hand-crafted rules.
Rules can be designed based on domain-specific gazetteers~\cite{sekine2004definition,etzioni2005unsupervised} and syntactic-lexical patterns~\cite{zhang2013unsupervised}.  Kim~\cite{kim2000rule} proposed to use Brill rule inference approach for speech input.
This system generates rules automatically based on Brill's part-of-speech tagger.
In biomedical domain,  Hanisch \etal~\cite{hanisch2005prominer} proposed ProMiner, which leverages a pre-processed synonym dictionary to identify protein mentions and potential gene in biomedical text.
Quimbaya \etal~\cite{quimbaya2016named} proposed a dictionary-based approach for NER in electronic health records.
Experimental results show the approach improves recall while having limited impact on precision.

Some other well-known rule-based NER systems include LaSIE-II~\cite{humphreys1998university}, NetOwl~\cite{krupka2005description}, Facile~\cite{black1998facile}, SAR~\cite{aone1998sra}, FASTUS~\cite{appelt1995sri}, and LTG~\cite{mikheev1999named} systems.
These systems are mainly based on hand-crafted semantic and syntactic rules to recognize entities.
Rule-based systems work very well when lexicon is exhaustive.
Due to domain-specific rules and incomplete dictionaries, high precision and low recall are often observed from such systems, and the systems cannot be transferred to other domains.

\subsubsection{Unsupervised Learning Approaches}

A typical approach of unsupervised learning is clustering~\cite{nadeau2007survey}. Clustering-based NER systems extract named entities from the clustered groups based on context similarity. The key idea is that lexical resources, lexical patterns, and statistics computed on a large corpus can be used to infer mentions of named entities.  Collins \etal~\cite{collins1999unsupervised} observed that use of unlabeled data reduces the requirements for supervision to just 7 simple ``seed'' rules.  The authors then presented two unsupervised algorithms for named entity classification.
Similarly,  KNOWITALL~\cite{etzioni2005unsupervised}  leveraged a set of predicate names as input and bootstraps its recognition process from a small set of generic extraction patterns.

Nadeau \etal~\cite{nadeau2006unsupervised} proposed an unsupervised system for gazetteer building and named entity ambiguity resolution.
This system combines entity extraction and disambiguation based on simple yet highly effective heuristics.
In addition, Zhang and Elhadad~\cite{zhang2013unsupervised} proposed an unsupervised approach to extracting named entities from biomedical text.
Instead of supervision, their model resorts to terminologies, corpus statistics (\eg inverse document frequency and context vectors) and shallow syntactic knowledge (\eg noun phrase chunking). Experiments on two mainstream biomedical datasets demonstrate the effectiveness and generalizability of their unsupervised approach.

\subsubsection{Feature-based Supervised Learning Approaches}
\label{featuedengineeredappraoches}

Applying supervised learning, NER is cast to a multi-class classification or sequence labeling task.
Given annotated data samples, features are carefully designed to represent each training example.
Machine learning algorithms are then utilized to learn a model to recognize similar patterns from unseen data.

Feature engineering is critical in supervised NER systems. Feature vector representation is an abstraction over text where a word is represented by one or many Boolean, numeric, or nominal values~\cite{sekine2009named,nadeau2007survey}.
Word-level features (\eg case, morphology, and part-of-speech tag)~\cite{zhou2002named,settles2004biomedical,liao2009simple}, list lookup features (\eg Wikipedia gazetteer and DBpedia gazetteer)~\cite{mikheev1999knowledge,kazama2007exploiting,toral2006proposal,hoffart2011robust}, and document and corpus features (\eg local syntax and multiple occurrences)~\cite{ravin1997extracting,zhu2005espotter,ji2016joint,krishnan2006effective} have been widely used in various supervised NER systems. More feature designs are discussed  in~\cite{nadeau2007survey,campos2012biomedical,sharnagat2014named}

Based on these features, many machine learning algorithms have been applied in supervised NER, including
Hidden Markov Models (HMM)~\cite{eddy1996hidden}, Decision Trees~\cite{quinlan1986induction},  Maximum Entropy Models~\cite{kapur1989maximum}, Support Vector Machines (SVM)~\cite{hearst1998support}, and Conditional Random
Fields (CRF)~\cite{lafferty2001conditional}.

Bikel \etal~\cite{bikel1997nymble,bikel1999algorithm} proposed the first HMM-based NER system, named IdentiFinder, to identify and classify names, dates, time expressions, and numerical quantities.
 In addition, Szarvas \etal~\cite{szarvas2006multilingual} developed a multilingual NER system by using  C4.5 decision tree and AdaBoostM1 learning algorithm.
 A major merit is that it provides an opportunity to train several independent decision tree classifiers through different subsets of features then combine their decisions through a majority voting scheme.
Borthwick \etal~\cite{borthwick1998nyu} proposed  ``maximum entropy named entity'' (MENE) by applying the maximum entropy theory.  MENE is able to make use of an extraordinarily diverse range of knowledge sources in making its tagging decisions.  Other systems using maximum entropy can be found in~\cite{bender2003maximum,chieu2002named,curran2003language}.

McNamee and Mayfield~\cite{mcnamee2002entity} used 1000 language-related and 258 orthography and punctuation features to train  SVM classifiers. Each classifier makes binary decision whether the current token belongs to one of the eight classes, \ie  B- (Beginning), I- (Inside) for \textit{PERSON}, \textit{ORGANIZATION}, \textit{LOCATION}, and \textit{MIS} tags.
SVM does not consider ``neighboring'' words when predicting an entity label. CRFs takes context into account.
McCallum and Li~\cite{mccallum2003early} proposed a feature induction method for CRFs in NER.
Experiments were performed on CoNLL03, and achieved F-score of 84.04\% for English.
Krishnan and Manning~\cite{krishnan2006effective} proposed a two-stage approach based on two coupled CRF classifiers.
The second CRF makes use of the latent representations derived from the output of the first CRF.
We note that CRF-based NER has been widely applied to texts in various domains, including biomedical text~\cite{settles2004biomedical,liu2019hamner}, tweets~\cite{ritter2011named,liu2011recognizing} and chemical text~\cite{rocktaschel2012chemspot}.

\section{Deep Learning Techniques for NER}
\label{sec:dlner}

In recent years, DL-based NER models become dominant and achieve state-of-the-art results.
Compared to feature-based approaches, deep learning is beneficial in discovering hidden features automatically.
Next, we first briefly introduce what deep learning is, and why deep learning for NER. We then survey DL-based NER approaches.

\subsection{Why Deep Learning for NER?}

Deep learning is a field of machine learning that is composed of multiple processing layers to learn representations of data with multiple levels of abstraction~\cite{lecun2015deep}. The typical layers are artificial neural networks which consists of the forward pass and backward pass. 
The forward pass computes a weighted sum of their inputs from the previous layer and pass the result through a non-linear function.
The backward pass is to compute the gradient of an objective function with respect to the weights of a multilayer stack
of modules via the chain rule of derivatives.
The key advantage of deep learning is the capability of representation learning and the semantic composition empowered by both the vector representation and neural processing. This  allows a machine to be fed with raw data and to automatically discover latent representations and processing needed for classification or detection \cite{lecun2015deep}.

There are three core strengths of applying deep learning techniques to NER. First, NER benefits from the non-linear transformation, which generates non-linear mappings from input to output. Compared with linear models (\eg log-linear HMM and linear chain CRF), DL-based models are able to learn complex and intricate features from data via non-linear activation functions.  Second, deep learning saves significant effort on designing NER features. The traditional feature-based approaches require considerable amount of engineering skill and domain expertise.
DL-based models, on the other hand, are effective in automatically learning useful representations and underlying factors from raw data. Third, deep neural NER models can be trained in an end-to-end paradigm, by gradient descent.
This property enables us to design possibly complex NER systems.

\begin{figure}[t]
	\centering	
	\includegraphics[width=0.9\columnwidth,draft=false]{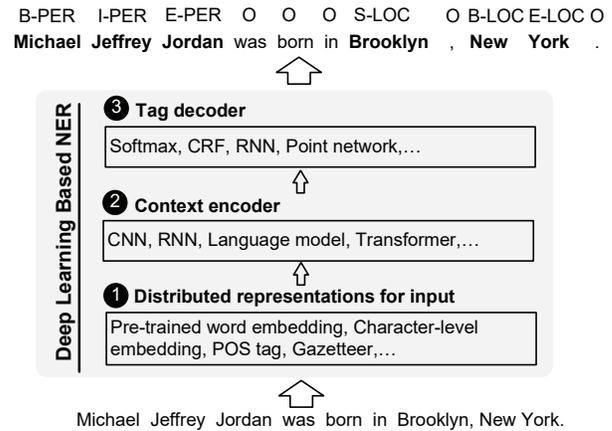}	
	\caption{The taxonomy of DL-based NER. From input sequence to predicted tags, a DL-based NER model consists of distributed representations for input, context encoder, and tag decoder.}
	\label{fig:deepner}
\end{figure}

\begin{figure*}[t]
	\centering
	\subfigure[CNN-based character-level representation.] {\includegraphics[height=1.5in,draft=false]{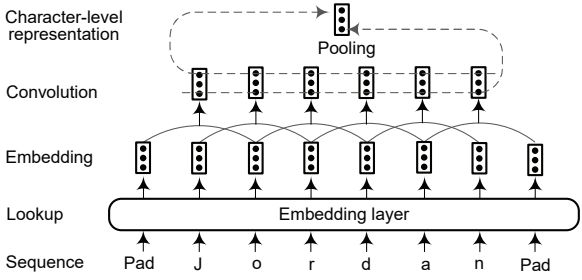}\label{fig:charCNN}}
	\hspace{0.5in}
	\subfigure[RNN-based character-level representation.] {\includegraphics[height=1.5in,draft=false]{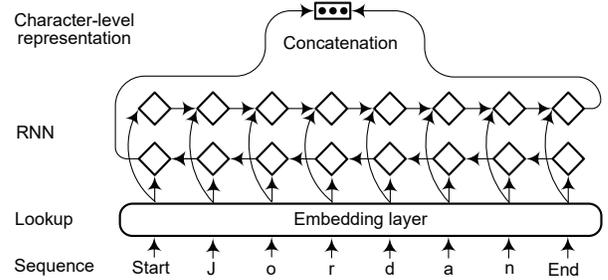}\label{fig:charRNN}}
	\vspace{-3mm}
	\caption{CNN-based and RNN-based models for extracting character-level representation for a word.}
\end{figure*}

\paratitle{Why we use a new taxonomy in this survey?} Existing taxonomy~\cite{shen2017deep,yadav2018survey} is based on character-level encoder, word-level encoder, and tag decoder.
We argue that the description of ``word-level encoder'' is inaccurate because word-level information is used twice in a typical DL-based NER model: 1)  word-level representations are used as raw features, and 2) word-level representations (together with character-level representations) are used to capture context dependence for tag decoding.
In this survey, we summarize recent advances in NER with the general architecture presented in Figure~\ref{fig:deepner}. \textit{Distributed representations for input} consider word- and character-level embeddings as well as incorporation of additional features like POS tag and gazetteer that have been effective in feature-based based approaches. \textit{Context encoder} is  to capture the context dependencies using CNN, RNN, or other networks. \textit{Tag decoder} predict tags for tokens in the input sequence. For instance, in Figure~\ref{fig:deepner} each token is predicted with a tag indicated by B-(begin), I-(inside), E-(end), S-(singleton) of a named entity with its type, or O-(outside) of named entities. Note that there are other tag schemes or tag notations, \eg BIO. Tag decoder may also be trained to detect entity boundaries and then the detected text spans are classified to the entity types.

\subsection{Distributed Representations for Input}
\label{ssec:input}

A straightforward option of representing a word is \textit{one-hot} vector representation.
In one-hot vector space, two distinct words have completely different representations and are orthogonal.
\textit{Distributed representation} represents words in low dimensional real-valued dense vectors where each dimension represents a latent feature.
Automatically learned from text, distributed representation captures semantic and syntactic properties of word, which do not explicitly present in the input to NER.  Next, we review three types of distributed representations that have been used in NER models: word-level, character-level, and hybrid representations.

\subsubsection{Word-level Representation}

Some studies~\cite{nguyen2016toward,zheng2017joint,strubell2017fast} employed word-level representation, which is typically pre-trained over large collections of text through unsupervised algorithms such as continuous bag-of-words (CBOW)  and  continuous skip-gram models~\cite{mikolov2013efficient}.
Recent studies~\cite{shen2017deep,yang2018design} have shown the importance of such pre-trained word embeddings. Using as the input, the pre-trained word embeddings can be either fixed or further fine-tuned during NER model training.
Commonly used word embeddings include Google Word2Vec, Stanford GloVe, Facebook  fastText and SENNA.

Yao \etal~\cite{yao2015biomedical} proposed Bio-NER, a biomedical NER model based on deep neural network architecture.
The word representation in Bio-NER is trained on PubMed database using skip-gram model.
The dictionary contains  205,924 words in 600 dimensional vectors.
Nguyen \etal~\cite{nguyen2016toward} used word2vec toolkit to learn word embeddings for English from the Gigaword corpus augmented with newsgroups data from BOLT (Broad Operational Language Technologies).
Zhai \etal~\cite{zhai2017neural} designed a neural model for sequence chunking, which consists of two sub-tasks: segmentation and labeling. The neural model can be fed with SENNA embeddings or randomly initialized embeddings.

Zheng \etal~\cite{zheng2017joint}  jointly extracted entities and relations using a single model.
This end-to-end model uses word embeddings learned on NYT corpus by  word2vec tookit.
Strubell \etal~\cite{strubell2017fast} proposed a tagging scheme based on Iterated Dilated Convolutional Neural Networks (ID-CNNs).
The lookup table in their model are initialized by 100-dimensional embeddings trained on SENNA corpus by skip-n-gram.
In their proposed neural model for extracting entities and their relations, Zhou  \etal~\cite{zhou2017joint} used the pre-trained 300-dimensional word vectors from Google. In addition,  GloVe~\cite{ma2016end, li2017leveraging} and  fastText~\cite{wang2018code} are also widely used in NER tasks.

\subsubsection{Character-level Representation}

Instead of only considering word-level representations as the basic input, several studies~\cite{kuru2016charner,tran2017named} incorporated character-based word representations learned from an end-to-end neural model.
Character-level representation has been found useful for exploiting explicit sub-word-level information such as prefix and suffix.
Another advantage of character-level representation is that it naturally handles out-of-vocabulary.
Thus character-based model is able to infer representations for unseen words and share information of morpheme-level regularities.
There are two widely-used architectures for extracting character-level representation: \textit{CNN-based} and \textit{RNN-based} models.
Figures~\ref{fig:charCNN} and~\ref{fig:charRNN} illustrate the two architectures.

Ma \etal~\cite{ma2016end} utilized a CNN for extracting character-level representations of words.
Then the character representation vector is concatenated with the word embedding before feeding into a RNN context encoder.
Likewise,  Li \etal~\cite{li2017leveraging}  applied a series of convolutional and highway layers to generate character-level representations for words.
The final embeddings of words are fed into a bidirectional recursive network.
Yang \etal~\cite{yang2017neural} proposed a neural reranking model for NER, where a convolutional
layer with a fixed window-size is used on top of a character embedding layer.
Recently, Peters \etal~\cite{peters2018deep} proposed ELMo word representation, which are computed on top of two-layer bidirectional language models with character convolutions.

For \textit{RNN-based} models, Long Short-Term Memory (LSTM) and Gated Recurrent Unit (GRU) are two typical choices of the basic units.
Kuru \etal~\cite{kuru2016charner} proposed CharNER, a character-level tagger for language-independent NER.
CharNER considers a sentence as a sequence of characters and utilizes LSTMs to extract character-level representations.
It outputs a tag distribution for each character instead of each word.
Then word-level tags are obtained from the character-level tags.
Their results show that taking characters as the primary representation is superior to words as the basic input unit.
Lample \etal~\cite{lample2016neural} utilized a bidirectional LSTM to extract character-level representations of words.
Similar to~\cite{ma2016end}, character-level representation is concatenated with pre-trained word-level embedding from a word lookup table.
Gridach~\cite{gridach2017character}  investigated word embeddings and character-level representation in identifying biomedical named entities.
Rei \etal~\cite{rei2016attending} combined character-level representations with word embeddings using a gating mechanism.
In this way, Rei's model  dynamically decides how much information to use from a character- or word-level component.
Tran \etal~\cite{tran2017named} introduced a neural NER model with stack residual LSTM and trainable bias decoding, where word features are extracted from word embeddings and character-level RNN.
Yang \etal~\cite{yang2016multi} developed a model to handle both cross-lingual and multi-task joint training in a unified manner.
They employed a deep bidirectional GRU to learn informative morphological representation from the character sequence of a word.
Then character-level representation  and word embedding are concatenated to produce the final representation for a word.

\begin{figure}[t]
	\centering	
	\includegraphics[width=0.98\columnwidth,draft=false]{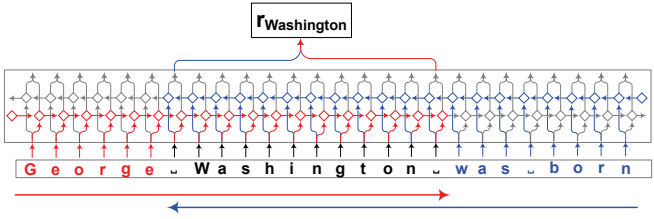}	
	\caption{Extraction of a contextual string embedding for  word ``Washington'' in a sentential context~\cite{akbik2018contextual}. From the forward language model (shown in red), the model extracts the output hidden state after the last character in the word. From the backward language model (shown in blue), the model extracts the output hidden state before the first character in the word. Both output hidden states are concatenated to form the final embedding of a word.}
	\label{fig:contextualstring}
\end{figure}

Recent advances in language modeling using recurrent neural networks made it viable to model language as distributions over characters.
 The contextual string embeddings by Akbik \etal~\cite{akbik2018contextual}, uses character-level neural language model to generate  a contextualized embedding for a string of characters in a sentential context.
An important property is that the embeddings are contextualized by their surrounding text, meaning that the same word has different embeddings depending on its contextual use.
Figure~\ref{fig:contextualstring} illustrates the architecture of extracting a contextual string embedding for  word ``Washington'' in a sentential context.

\subsubsection{Hybrid Representation}

Besides word-level and character-level representations, some studies also incorporate additional information (\eg gazetteers~\cite{huang2015bidirectional,DBLP:conf/acl/LiuYL19}, lexical similarity~\cite{ghaddar2018robust},  linguistic dependency \cite{DBLP:journals/corr/abs-1909-10148} and visual features \cite{lu2018visual}) into the final representations of words, before feeding into context encoding layers. In other words, the DL-based representation is combined with feature-based approach in a hybrid manner.
Adding additional information may lead to  improvements in NER performance, with the price of hurting generality of these systems.

The use of neural models for NER was pioneered by~\cite{collobert2011natural}, where an architecture based on temporal convolutional neural networks over word sequence was proposed.
When incorporating common priori knowledge (\eg gazetteers and POS), the resulting system outperforms the baseline using only word-level representations.
In the BiLSTM-CRF model by Huang \etal~\cite{huang2015bidirectional},  four types of features are used for the NER task: spelling features, context features, word embeddings, and gazetteer features.
Their experimental results show that the extra features (\ie gazetteers) boost tagging accuracy.
The BiLSTM-CNN model by Chiu and Nichols~\cite{chiu2016named} incorporates a bidirectional LSTM and a character-level CNN.
Besides word embeddings, the model uses additional word-level features (capitalization, lexicons)
 and character-level features (4-dimensional vector representing the type of a character: upper case, lower case, punctuation, other).

Wei \etal~\cite{wei2016disease} presented a CRF-based neural system for recognizing and normalizing disease names.
 This system employs rich features in addition to word embeddings, including words, POS tags, chunking, and word shape features (\eg dictionary
 and morphological features).
Strubell \etal~\cite{strubell2017fast} concatenated 100-dimensional embeddings with a 5-dimensional word shape vector (\eg  all capitalized, not capitalized, first-letter capitalized or contains a capital letter).
Lin \etal~\cite{lin2017multi} concatenated character-level representation, word-level representation, and syntactical word representation (\ie POS tags, dependency roles, word positions, head positions) to form a comprehensive word representation.
A multi-task approach for NER was proposed by Aguilar \etal~\cite{aguilar2017multi}.
This approach utilizes a CNN to capture orthographic features and word shapes at character level.
For syntactical and contextual information at word level, \eg POS and word embeddings, the model implements a LSTM architecture.
 Jansson and Liu~\cite{jansson2017distributed} proposed to combine Latent Dirichlet Allocation (LDA) with  deep learning on character-level and word-level embeddings.

Xu \etal~\cite{xu2017local} proposed a local detection approach for NER based on fixed-size ordinally forgetting encoding (FOFE)~\cite{zhang2015fixed}, FOFE explores both character-level and word-level representations for each fragment and its contexts.
In the multi-modal NER system by Moon \etal~\cite{moon2018multimodal}, for noisy user-generated data like tweets and Snapchat captions,
word embeddings, character embeddings, and visual features are merged with modality attention.
Ghaddar and Langlais~\cite{ghaddar2018robust} found that it was unfair that lexical features had been mostly discarded in neural NER systems.
They proposed an alternative lexical representation which is trained offline and can be added to any neural NER system.
The lexical representation is computed for each word with a 120-dimensional vector, where each element encodes the
similarity of the word with an entity type.
Recently, Devlin \etal~\cite{devlin2018bert} proposed a new language representation model called BERT, bidirectional encoder representations from transformers.
BERT uses masked language models to enable pre-trained deep bidirectional representations.
For a given token, its input representation is comprised by summing the corresponding position, segment and token embeddings.
Note that pre-trained language model embeddings often require large-scale corpora for training, and intrinsically incorporate auxiliary embeddings (\eg position and segment embeddings). 
For this reason, we category these contextualized language-model embeddings as hybrid representations in this survey.

\begin{figure}[t]
	\centering	
	\includegraphics[width=0.9\columnwidth,draft=false]{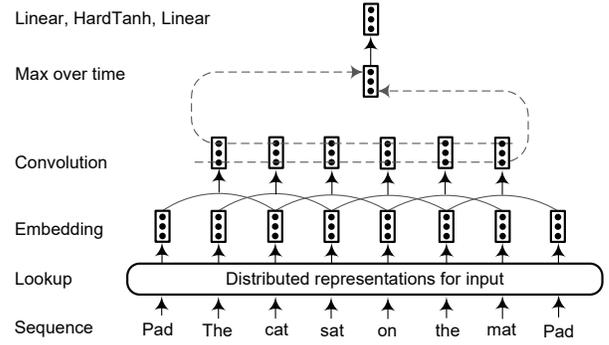}	
	\caption{Sentence approach network based on CNN~\cite{collobert2011natural}. The convolution layer extracts features from the whole sentence, treating it as a \textit{sequence} with global structure.	}
	\label{fig:wordCNN}
	\vspace{-5mm}
\end{figure}
\subsection{Context Encoder Architectures}
\label{ssec:contextEncoder}
Here, we now review widely-used context encoder architectures: convolutional neural networks, recurrent neural networks, recursive neural networks, and deep transformer.

\subsubsection{Convolutional Neural Networks}

Collobert \etal~\cite{collobert2011natural} proposed a sentence approach network where a word is tagged with the consideration of  \textit{whole} sentence, shown in Figure~\ref{fig:wordCNN}.
Each word in the input sequence is embedded to an $N$-dimensional vector after the stage of input representation.
Then a convolutional layer is used to produce local features around each word, and
the size of the output of the convolutional layers depends on the number of words in the sentence.
The global feature vector is constructed by combining local feature vectors extracted by the convolutional layers.
The dimension of the global feature vector is fixed, independent of the sentence length, in order to apply subsequent standard affine layers.
Two approaches are widely used to extract global features: a max or an averaging operation over the position (\ie ``time'' step) in the sentence.
Finally, these fixed-size global features are fed into tag decoder to compute distribution scores for all possible tags for the words in the network input.
Following Collobert's work, Yao \etal~\cite{yao2015biomedical} proposed Bio-NER for biomedical NER.
Wu \etal~\cite{wu2015named} utilized a convolutional layer to generate  global features represented by a number of global hidden nodes.
Both local features and global features are then fed into a standard affine network to recognize named entities in clinical text.

Zhou \etal~\cite{zhou2017joint} observed that with RNN latter words influence the final sentence representation more than former words. However, important words may appear anywhere in a sentence.
In their proposed model, named  BLSTM-RE, BLSTM  is used to capture long-term dependencies and obtain the whole representation of an input sequence.
CNN is then utilized to learn a high-level representation, which is then fed into a sigmoid classifier.
Finally, the whole sentence representation (generated by BLSTM) and the relation presentation (generated by the sigmoid classifier) are fed into another LSTM to predict entities.

\begin{figure}[t]
	\centering	
	\includegraphics[width=0.9\columnwidth,draft=false]{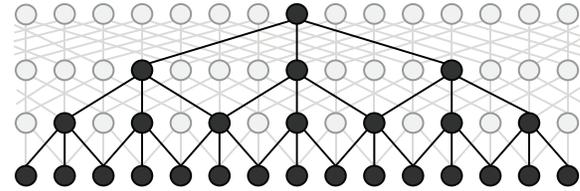}	
	\caption{The architecture of ID-CNNs with filter width 3 and maximum dilation width 4.~\cite{strubell2017fast}.
	}
	\label{fig:idcnn}
\end{figure}

Traditionally, the time complexity of LSTMs for a sequence of length $N$ is $\mathcal{O(N)}$ in a parallelism manner. 
 Strubell \etal~\cite{strubell2017fast} proposed ID-CNNs, referred to Iterated Dilated Convolutional Neural Networks, which is more computationally efficient due to the capacity of handling larger context and structured prediction. 
 Figure~\ref{fig:idcnn} shows the architecture of a dilated CNN block, where four stacked dilated convolutions of width 3 produce token representations. 
Experimental results show that ID-CNNs achieves 14-20x test-time speedups compared to Bi-LSTM-CRF while retaining comparable accuracy.

\subsubsection{Recurrent Neural Networks}

\begin{figure}[t]
	\centering	
	\includegraphics[width=0.9\columnwidth,draft=false]{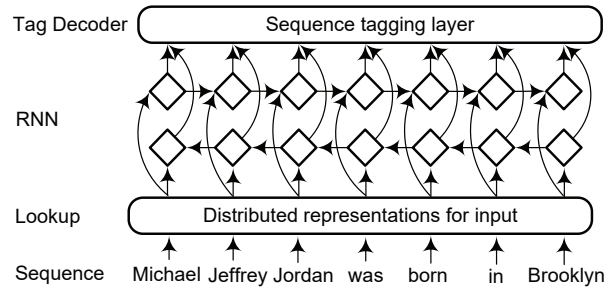}	
	\caption{The architecture of RNN-based context encoder.
	}
	\label{fig:wordRNN}
\end{figure}

Recurrent neural networks, together with its variants such as gated recurrent unit (GRU) and  long-short term memory (LSTM), have demonstrated remarkable achievements in modeling sequential data.
In particular,  bidirectional RNNs efficiently make use of past information (via forward states) and future information (via backward states) for a specific time frame \cite{huang2015bidirectional}.
Thus, a token encoded by a bidirectional RNN will contain evidence from the whole input sentence.
Bidirectional RNNs therefore become de facto standard for composing deep context-dependent representations of text~\cite{strubell2017fast,ma2016end}.
A typical architecture of RNN-based context encoder is shown in Figure~\ref{fig:wordRNN}.

The work by Huang \etal~\cite{huang2015bidirectional} is among the first to utilize a bidirectional LSTM CRF architecture to sequence tagging tasks (POS, chunking and NER).
Following~\cite{huang2015bidirectional}, a body of works~\cite{ma2016end,chiu2016named,rei2016attending,wei2016disease,lample2016neural,nguyen2016toward,zhai2017neural,zheng2017joint,tran2017named,zhou2017joint,lin2017multi} applied BiLSTM as the basic architecture to encode sequence context information.
Yang \etal~\cite{yang2016multi} employed deep GRUs on both character and word levels to encode morphology and context information.
They further extended their model to cross-lingual  and multi-task joint trained by sharing the architecture and parameters.

Gregoric \etal~\cite{gregoric2018named} employed multiple independent bidirectional LSTM units across the same input. Their model promotes diversity among the LSTM units by employing an inter-model regularization term.
By distributing computation across multiple smaller LSTMs, they found a reduction in total number of parameters.
Recently, some studies~\cite{katiyar2018nested,ju2018neural} designed LSTM-based neural networks for nested named entity recognition.
Katiyar and Cardie~\cite{katiyar2018nested} presented a modification to standard LSTM-based sequence labeling model to handle nested named entity recognition.
Ju \etal~\cite{ju2018neural} proposed a neural model to identify nested entities by dynamically stacking flat NER layers until no outer entities are extracted.
Each flat NER layer employs bidirectional LSTM to capture sequential context.
The model merges the outputs of the LSTM layer in the current flat NER layer to construct new representations for the detected entities and then feeds them into the next flat NER layer.

\subsubsection{Recursive Neural Networks}

\begin{figure}[t]
	\centering	
	\includegraphics[width=1\columnwidth,draft=false]{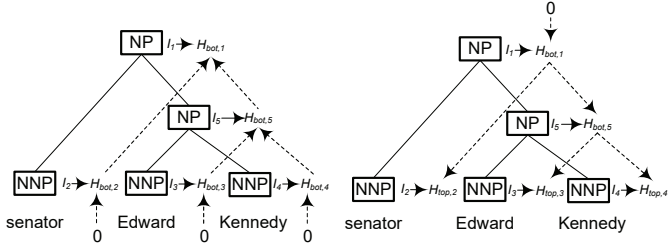}	
	\caption{Bidirectional recursive neural networks for NER~\cite{li2017leveraging}. The computations are done recursively in two directions. The
			bottom-up direction computes the semantic composition of the subtree of each node, and the top-down counterpart propagates to that node the linguistic structures which contain the subtree.	}
	\label{fig:wordRecursive}
\end{figure}

Recursive neural networks are non-linear adaptive models that are able to learn deep structured information, by traversing a given structure in topological order.
Named entities are highly related to linguistic constituents, \eg noun phrases~\cite{li2017leveraging}.
However, typical sequential labeling approaches take little into consideration about phrase structures of sentences.
To this end, Li \etal \cite{li2017leveraging} proposed to classify every node in a constituency structure for NER.
This model recursively calculates hidden state vectors of every node and classifies each node by these hidden vectors.
Figure~\ref{fig:wordRecursive} shows how to recursively compute two hidden state features for every node.
The bottom-up direction calculates the semantic composition of the subtree of each node, and the top-down counterpart propagates to that node the linguistic structures which contain the subtree.
Given hidden vectors for every node, the network calculates a probability distribution of entity types plus a special non-entity type.

\begin{figure}[t]
	\centering	
	\includegraphics[width=1\columnwidth,draft=false]{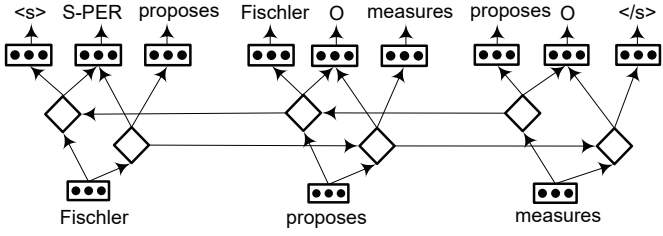}	
	\caption{A sequence labeling model with an additional language modeling objective~\cite{rei2017semi}, performing NER on the sentence ``\textit{Fischler proposes measures}''. At each token position (\eg ``proposes''), the network is optimised to predict the previous word (``Fischler''), the current label (``O''), and the next word (``measures'') in the sequence.
	}
	\label{fig:reiLM}
\end{figure}

\subsubsection{Neural Language Models}

Language model is a family of models describing the generation of sequences.
Given a token sequence, $(t_1,t_2,\ldots,t_N)$, a forward language model computes the probability of the sequence by modeling the probability of token $t_k$ given its history $(t_1,\ldots,t_{k-1})$ \cite{peters2017semi}:
\begin{equation}
p(t_1,t_2,\ldots,t_N)=\prod\limits_{k = 1}^N {p(t_k|t_1,t_2,\ldots,t_{k-1})}
\end{equation}
A backward language model is similar to a forward language model, except it runs over the sequence in reverse order, predicting the previous token given its future context:
\begin{equation}
p(t_1,t_2,\ldots,t_N)=\prod\limits_{k = 1}^N {p({t_k}|{t_{k+1},t_{k+2},...,t_{N}})}
\end{equation}
For neural language models, probability of token $t_k$ can be computed by the output of recurrent neural networks. At each position $k$, we can obtain two context-dependent representations (forward and backward) and then combine them as the final language model embedding for token $t_k$.
Such language-model-augmented knowledge has been empirically verified to be helpful in numerous sequence labeling tasks~\cite{rei2017semi,peters2017semi,liu2018efficient,liu2017empower,peters2018deep,DBLP:conf/acl/JiaXZ19}.

Rei~\cite{rei2017semi}  proposed a framework with a secondary objective -- learning to predict surrounding words for each word in the dataset.
Figure~\ref{fig:reiLM} illustrates the architecture with a short sentence on the NER task.
At each time step (\ie token position), the network is optimised to predict the previous token, the current tag, and the next token in the sequence.
The added language modeling objective encourages the system to learn richer feature representations which are then reused for sequence labeling.

\begin{figure}[t]
	\centering	
	\includegraphics[width=1\columnwidth,draft=false]{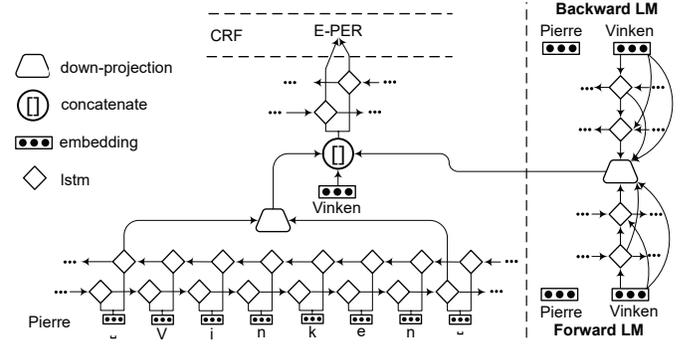}	
	\caption{Sequence labeling architecture with contextualized representations~\cite{liu2018efficient}. Character-level representation,  pre-trained word embedding and contextualized representation from bidirectional language models are concatenated and further fed into context encoder.
	}
	\label{fig:liuLM}
\end{figure}

Peters \etal~\cite{peters2017semi} proposed TagLM, a language model augmented sequence tagger. This tagger considers both  pre-trained word embeddings and bidirectional language model embeddings for every token in the input sequence for sequence labeling task.
Figure~\ref{fig:liuLM} shows the architecture of  LM-LSTM-CRF model ~\cite{liu2017empower,liu2018efficient}. The language model and sequence tagging model share the same character-level layer in a multi-task learning manner. The vectors from character-level embeddings, pre-trained word embeddings, and language model representations, are concatenated and fed into the word-level LSTMs.
Experimental results demonstrate that multi-task learning is an effective approach to guide the language model to learn task-specific knowledge.

\begin{figure*}[t]
	\centering
	\subfigure[Google BERT] {\includegraphics[height=1.45in,draft=false]{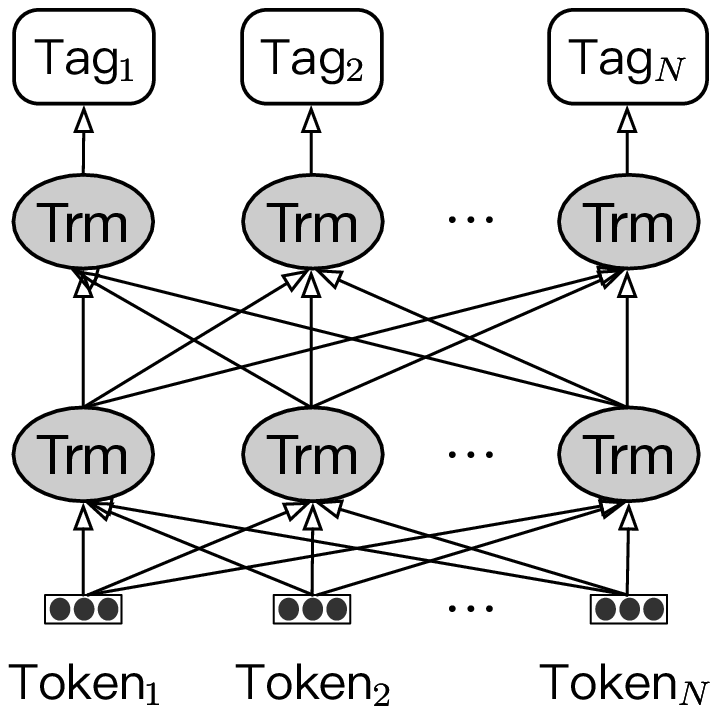}}
	\hfill
	\subfigure[OpenAI GPT] {\includegraphics[height=1.45in,draft=false]{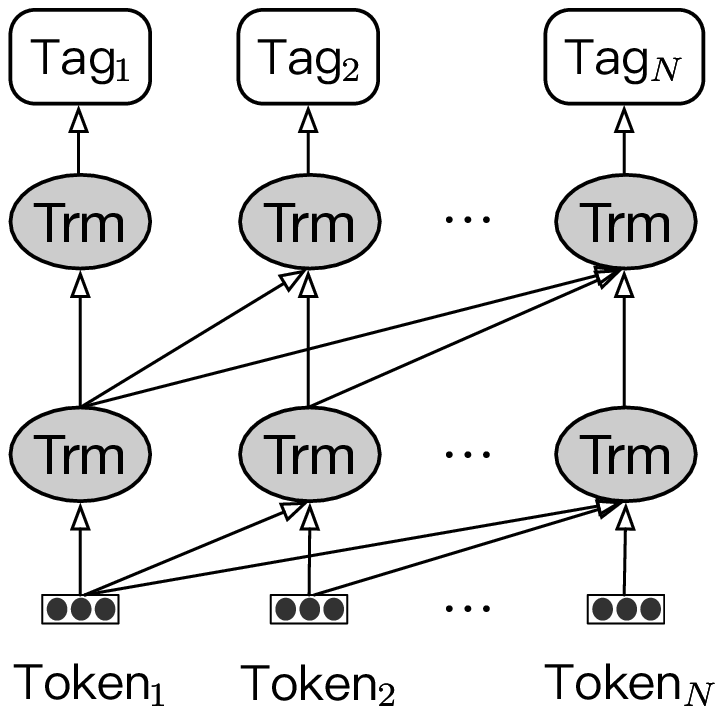}}	\hfill
	\subfigure[AllenNLP ELMo] {\includegraphics[height=1.45in,draft=false]{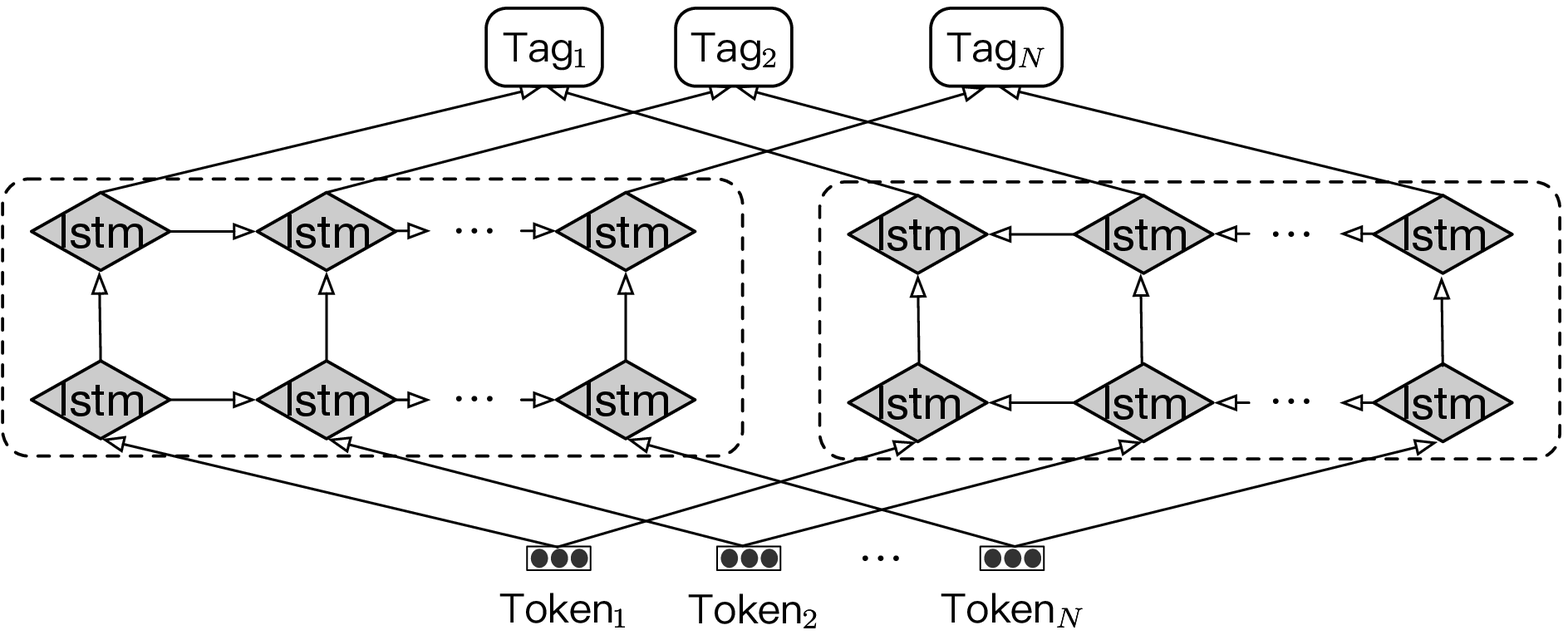}}
	\caption{Differences in pre-training model architectures~\cite{devlin2018bert}. Google BERT uses a bidirectional Transformer (abbreviated as ``Trm''). OpenAI GPT uses a left-to-right Transformer. AllenNLP ELMo uses the concatenation of independently trained left-to-right and right-to-left LSTM to generate features for downstream tasks.	}
	\label{fig:bertGPTElmo}
\end{figure*}

Figure~\ref{fig:contextualstring} shows the contextual string embedding using neural character-level language modeling by Akbik \etal~\cite{akbik2018contextual}.
They utilized the hidden states of a forward-backward recurrent neural network to create contextualized word embeddings.
A major merit of this model is that character-level language model is independent of tokenization and a fixed vocabulary.
Peters \etal~\cite{peters2018deep} proposed ELMo representations, which are computed on top of two-layer bidirectional language models with character convolutions.
This new type of deep contextualized word representation is capable of modeling both complex characteristics of word usage (\eg semantics and syntax), and usage variations across linguistic contexts (\eg polysemy).

\subsubsection{Deep Transformer}
\label{sssec:transformer}

Neural sequence labeling models are typically based on complex convolutional or recurrent networks which consists of encoders and decoders.
\textit{Transformer}, proposed by Vaswani \etal~\cite{vaswani2017attention},  dispenses with recurrence and convolutions entirely. 
Transformer utilizes stacked self-attention and point-wise, fully connected layers to build basic blocks for encoder and decoder. Experiments on various tasks~\cite{vaswani2017attention,liu2018generating,kitaev2018constituency} show Transformers to be superior in quality while requiring significantly less time to train.

Based on transformer, Radford \etal~\cite{radford2018improving} proposed Generative Pre-trained Transformer (GPT) for language understanding tasks.
GPT has a two-stage training procedure. First, they use a language modeling objective with Transformers on  unlabeled data to learn the initial parameters.
Then they adapt these parameters to a target task using the supervised objective, resulting in minimal changes to the pre-trained model.
Unlike GPT (a left-to-right architecture), Bidirectional Encoder Representations from Transformers (BERT) is proposed to pre-train deep bidirectional Transformer by jointly conditioning on both left and right context in all layers~\cite{devlin2018bert}.
Figure~\ref{fig:bertGPTElmo} summarizes BERT~\cite{devlin2018bert},  GPT~\cite{radford2018improving}  and  ELMo~\cite{peters2018deep}.
In addition, Baevski \etal \cite{DBLP:journals/corr/abs-1903-07785} proposed a novel cloze-driven pre-training regime based on a
bi-directional Transformer, which is trained with a cloze-style objective and predicts the center word given all left and right context.

These language model embeddings pre-trained using Transformer are becoming a new paradigm of NER. 
First, these embeddings are contextualized and can be used to replace the traditional embeddings, such as Google Word2vec and Stanford GloVe. 
Some studies \cite{DBLP:journals/corr/abs-1909-10148,DBLP:conf/acl/LiuYL19,DBLP:conf/acl/XiaZYLDWFMY19,DBLP:journals/corr/abs-1911-02257,DBLP:conf/acl/LiuMZXCZ19,jiang2019improved} have achieved promising performance via leveraging the combination of traditional embeddings and language model embeddings. 
Second, these language model embeddings can be further fine-tuned with one additional output layer for a wide range of tasks including NER and chunking.
Especially, Li \etal \cite{DBLP:journals/corr/abs-1910-11476,DBLP:journals/corr/abs-1911-02855} framed the NER task as a machine reading comprehension (MRC) problem, which can be solved by fine-tuning the BERT model.

\subsection{Tag Decoder Architectures}
\label{ssec:tagDecoder}
\begin{figure*}[t]
	\centering
	\subfigure[MLP+Softmax] {\includegraphics[height=1.5in,draft=false]{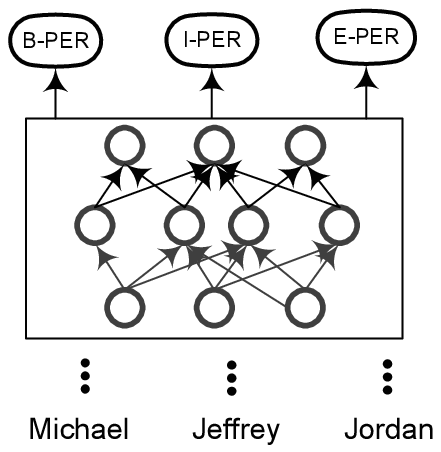}}
	\hfill
	\subfigure[CRF] {\includegraphics[height=1.5in,draft=false]{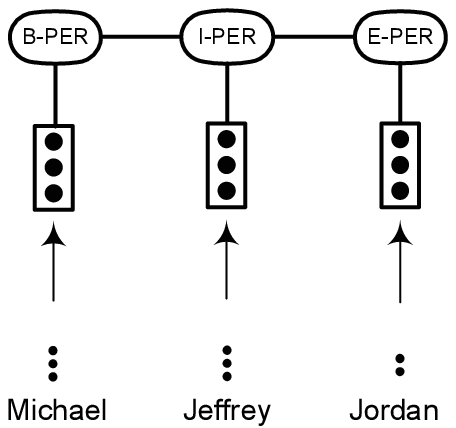}}	\hfill
	\subfigure[RNN] {\includegraphics[height=1.5in,draft=false]{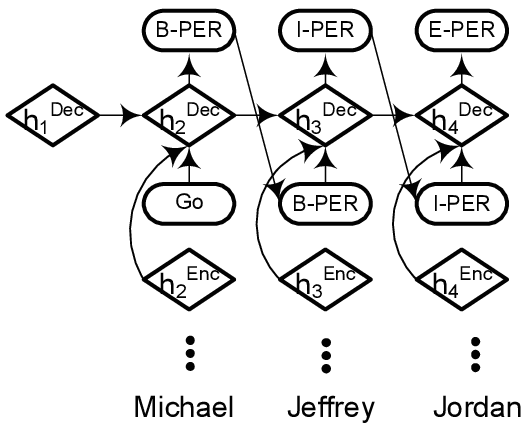}\label{sfig:tagrnn}} 	\newline
	\subfigure[Pointer Network] {\includegraphics[width=0.72\textwidth,draft=false]{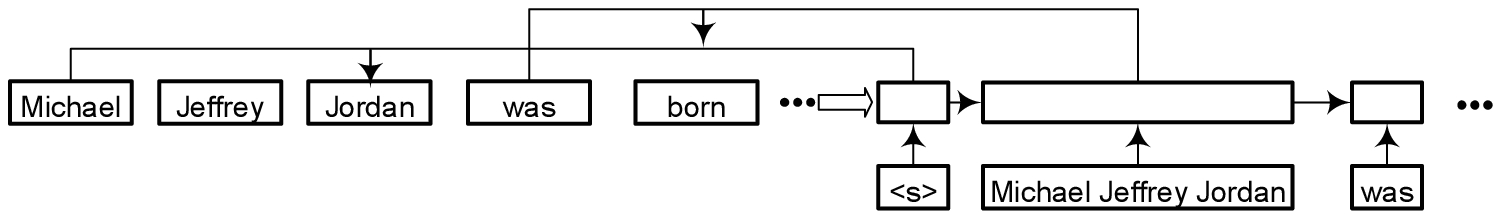}\label{sfig:pointer}}
\caption{Differences in four tag decoders: MLP+Softmax, CRF, RNN, and Pointer network.	}
\label{fig:tagDecoder}
\end{figure*}

Tag decoder is the final stage in a NER model. It takes context-dependent representations as input and produce a sequence of tags corresponding to the input sequence. Figure~\ref{fig:tagDecoder} summarizes four architectures of tag decoders: MLP + softmax layer,  conditional random fields (CRFs), recurrent neural networks, and pointer networks.

\subsubsection{Multi-Layer Perceptron + Softmax}
NER is in general formulated as a sequence labeling problem. With a multi-layer Perceptron + Softmax layer as the tag decoder layer, the sequence labeling task is cast as a multi-class classification problem. Tag for each word is independently predicted based on the context-dependent representations without taking into account its neighbors.

 A number of NER models~\cite{xu2017local,li2017leveraging,strubell2017fast,devlin2018bert,DBLP:journals/corr/abs-1908-08676} that have been introduced earlier use MLP + Softmax as the tag decoder. As a domain-specific NER task, Tomori \etal~\cite{tomori2016domain} used softmax as tag decoder to predict game states in Japanese chess game. Their model takes both input from text and input from chess board ($9\times 9$ squares with 40 pieces of 14 different types) and predict 21 named entities specific to this game.  Text representations and game state embeddings are both fed to a softmax layer for prediction of named entities using  BIO tag scheme.

\subsubsection{Conditional Random Fields}

A conditional random field (CRF) is a random field globally conditioned on the observation sequence \cite{lafferty2001conditional}.
CRFs have been widely used in feature-based supervised learning approaches (see Section~\ref{featuedengineeredappraoches}). Many deep learning based NER models use a CRF layer as the tag decoder, \eg on top of an bidirectional LSTM layer~\cite{zheng2017joint,peters2018deep,huang2015bidirectional,DBLP:conf/acl/LinLJYH19}, and on top of a CNN layer~\cite{yao2015biomedical,collobert2011natural,strubell2017fast}. Listed in Table~\ref{tab:tabsurvey}, CRF is the most common choice for tag decoder, and the state-of-the-art performance on CoNLL03 and OntoNotes5.0 is achieved by~\cite{akbik2018contextual} with a CRF tag decoder.

CRFs, however, cannot make full use of segment-level information because the inner properties of segments cannot be fully encoded with word-level representations.
Zhuo \etal~\cite{zhuo2016segment} then proposed gated recursive semi-markov CRFs, which directly model segments instead of words, and automatically extract segment-level features through a gated recursive convolutional neural network.
Recently, Ye and Ling~\cite{ye2018hybrid} proposed hybrid semi-Markov CRFs for neural sequence labeling.
This approach adopts segments instead of words as the basic units for feature extraction and transition modeling.
Word-level labels are utilized in deriving segment scores.
Thus, this approach is able to leverage both word- and segment-level information for segment score calculation.

\begin{table*}[t]
	\centering
	\caption{ Summary of recent works on neural NER. LSTM: long short-term memory, CNN: convolutional neural network, GRU: gated recurrent unit, LM: language model, ID-CNN: iterated dilated convolutional neural network, BRNN: bidirectional recursive neural network, MLP: multi-layer perceptron, CRF: conditional random field, Semi-CRF: Semi-markov conditional random field, FOFE: fixed-size ordinally forgetting encoding. }	
	\label{tab:tabsurvey}
		\scalebox{0.98}{
	\begin{tabular}{c|c@{}cp{35mm}|c|c|p{30mm}}
		\hline
		\multirow{2}{*}{Work} & \multicolumn{3}{c|}{Input representation} & \multirow{2}{*}{\tabincell{c}{Context  \\ encoder}} & \multirow{2}{*}{Tag decoder} & \multirow{2}{*}{Performance (F-score)} \\ \cline{2-4}
		& Character      & Word      & Hybrid      &                                  &                              &                              \\  \hline
		\cite{yao2015biomedical}&   -     & Trained on PubMed    &  POS         &          CNN              &   CRF                           &    GENIA: 71.01\%                          \\
		\cite{nguyen2016toward}&    -            &  Trained on  Gigaword        &  -          &         GRU                         &       GRU                       &    ACE 2005: 80.00\%                          \\
		\cite{zhai2017neural}&     -           &  Random         &   -         &          LSTM                        &    Pointer Network                          &      ATIS: 96.86\%                        \\
		\cite{zheng2017joint}&  -              &   Trained on NYT        &  -          &       LSTM                           &   LSTM            &         NYT: 49.50\%                     \\
		\cite{strubell2017fast} &    -            &  SENNA         & Word shape            &   ID-CNN                               &       CRF                       &     CoNLL03: 90.65\%;  \newline OntoNotes5.0: 86.84\%                         \\
		\cite{zhou2017joint}&       -         &  Google word2vec         &    -        &  LSTM                                & LSTM                       &      CoNLL04: 75.0\%                        \\
		\cite{kuru2016charner}	&    LSTM            &    -       &       -     &          LSTM                        &              CRF                &    CoNLL03: 84.52\%                     \\
		\cite{ma2016end} &        CNN        &     GloVe      &     -       &  LSTM                                &           CRF                   &      CoNLL03: 91.21\%                  \\
		\cite{rei2016attending}&      LSTM          &     Google word2vec      &   -         &      LSTM                            &        CRF                      &      CoNLL03: 84.09\%                  \\
		\cite{lample2016neural}	&     LSTM           &  SENNA         &    -        &    LSTM                              &           CRF                   &      CoNLL03: 90.94\%                  \\
		\cite{yang2016multi} 	&    GRU            &   SENNA        &    -        &        GRU                          &       CRF                       &    CoNLL03: 90.94\%                    \\
		\cite{li2017leveraging}&      CNN          &      GloVe     &    POS        &                     BRNN             &               Softmax               &         OntoNotes5.0: 87.21\%               \\
		\cite{akbik2018contextual} &       LSTM-LM         &     -      &        -    &           LSTM                       &               CRF               &            CoNLL03: 93.09\%; \newline OntoNotes5.0: 89.71\%        \\
		\cite{peters2018deep}& CNN-LSTM-LM             & -          &    -        &        LSTM                          &       CRF                       &    CoNLL03: 92.22\%                    \\
		\cite{collobert2011natural}&    -            &    Random       &   POS         &       CNN                           &    CRF                          &           CoNLL03: 89.86\%              \\
		\cite{huang2015bidirectional} &   -             &  SENNA         &     Spelling, n-gram, gazetteer       &    LSTM                               &            CRF                  &          CoNLL03: 90.10\%               \\
		\cite{chiu2016named}&    CNN            &   SENNA        & capitalization, lexicons           &  LSTM                                 &           CRF                   &      CoNLL03: 91.62\%; \newline OntoNotes5.0: 86.34\%                 \\
		\cite{xu2017local} &     -           &   -        &  FOFE          &          MLP                        &             CRF                 &    CoNLL03: 91.17\%                    \\
		\cite{tran2017named}&   LSTM             & GloVe           & -           &       LSTM                           &     CRF                         &       CoNLL03: 91.07\%                   \\
		\cite{lin2017multi}&       LSTM         & GloVe          & Syntactic            &        LSTM                          &       CRF                       &    W-NUT17: 40.42\%                     \\
		\cite{yang2017neural}&    CNN            &  SENNA         &  -          &      LSTM                            &    Reranker                          &       CoNLL03: 91.62\%                  \\
		\cite{aguilar2017multi}&    CNN            &   Twitter Word2vec        &   POS         &         LSTM                         &       CRF                       &     W-NUT17: 41.86\%                   \\
		\cite{jansson2017distributed}	&        LSTM        &  GloVe         &    POS, topics        &     LSTM                             &      CRF                        &           W-NUT17: 41.81\%              \\
		\cite{moon2018multimodal}	&     LSTM           &   GloVe        &     Images       &   LSTM                               &       CRF                       & SnapCaptions: 52.4\%                        \\
		\cite{ghaddar2018robust} 	&   LSTM             &    SSKIP       &     Lexical       &      LSTM                            &       CRF                       &   CoNLL03: 91.73\%;    \newline OntoNotes5.0: 87.95\%                    \\
		\cite{devlin2018bert} 	&     -           &  WordPiece         &     Segment, position       &         Transformer                         &                 Softmax             &           CoNLL03: 92.8\%             \\
		\cite{gregoric2018named}	&       LSTM         &     SENNA      &  -          &               LSTM                   &      Softmax                        &   CoNLL03: 91.48\%                       \\
		\cite{rei2017semi}	&      LSTM          & Google Word2vec          &   -         &        LSTM                          &        CRF                      &   CoNLL03: 86.26\%                         \\
		\cite{peters2017semi} 	&   GRU             &    SENNA       & LM           &               GRU                   &       CRF                       &    CoNLL03: 91.93\%                      \\
		\cite{liu2017empower}	&          LSTM      &     GloVe      &   -         &        LSTM                          &        CRF                      &        CoNLL03: 91.71\%                        \\
		\cite{zhuo2016segment} 		&     -           &    SENNA       &   POS, gazetteers         &         CNN                         &      Semi-CRF                        &       CoNLL03: 90.87\%                             \\
		\cite{ye2018hybrid}			& LSTM               &    GloVe       &   -         &                  LSTM                &     Semi-CRF                         &     CoNLL03: 91.38\%                           \\
		\cite{shen2017deep}				&      CNN          &        Trained on  Gigaword         &    -        &           LSTM                       &           LSTM                   &         CoNLL03: 90.69\%; \newline OntoNotes5.0: 86.15\%                \\ 
		\cite{DBLP:journals/corr/abs-1909-10148} &      -          &        GloVe         &  ELMo, dependency        &           LSTM                       &           CRF                   & CoNLL03: 92.4\%;\newline  OntoNotes5.0: 89.88\%                \\ 
		
		\cite{DBLP:conf/acl/LiuYL19}&      CNN         &        GloVe         &  ELMo, gazetteers        &           LSTM                       &           Semi-CRF                   & CoNLL03: 92.75\%; \newline OntoNotes5.0: 89.94\%              \\ 
		
			\cite{DBLP:conf/acl/XiaZYLDWFMY19}&      LSTM         &        GloVe         &  ELMo, POS        &           LSTM                       &           Softmax                   & CoNLL03: 92.28\%             \\ 
		
			\cite{DBLP:journals/corr/abs-1910-11476}&      -         &        -         & BERT       &           -                       &           Softmax                   &CoNLL03: 93.04\%; \newline OntoNotes5.0: 91.11\%          \\ 
			
		\cite{DBLP:journals/corr/abs-1911-02855}&      -         &        -         & BERT       &           -                       &           Softmax +Dice Loss                  &CoNLL03: 93.33\%; \newline \textbf{OntoNotes5.0: 92.07}\%          \\ 
		
		\cite{DBLP:journals/corr/abs-1911-02257}&      LSTM         &        GloVe         & BERT, document-level embeddings       &           LSTM                     &           CRF               &CoNLL03: 93.37\%; \newline OntoNotes5.0: 90.3\%          \\ 

		\cite{DBLP:conf/acl/LiuMZXCZ19}&      CNN         &        GloVe         & BERT, global embeddings       &           GRU                     &           GRU               &CoNLL03: 93.47\%          \\ 

\cite{DBLP:journals/corr/abs-1903-07785}&      CNN         &        -         &  Cloze-style LM embeddings    &           LSTM                     &           CRF               & \textbf{CoNLL03: 93.5\%}        \\ 

\cite{jiang2019improved}&      -         &        GloVe         &  Plooled contextual embeddings    &           RNN                     &           CRF               & CoNLL03: 93.47\%       \\

		 \hline
	\end{tabular}}
\end{table*}

\subsubsection{Recurrent Neural Networks}

A few studies~\cite{zheng2017joint,zhou2017joint,vaswani2016supertagging,nguyen2016toward,shen2017deep} have explored RNN to decode tags.
Shen \etal~\cite{shen2017deep} reported that RNN tag decoders outperform CRF and are faster to train when the number of entity types is large.
Figure~\ref{sfig:tagrnn} illustrates the workflow of RNN-based tag decoders, which serve as a language model to greedily produce a tag sequence.
The [GO]-symbol at the first step is provided as $y_1$ to the RNN decoder. Subsequently, at each time step $i$, the RNN decoder computes current decoder hidden state $h_{i+1}^{Dec}$ in terms of  previous step tag $y_i$,  previous step decoder hidden state $h_i^{Dec}$ and current step encoder hidden state $h_{i+1}^{Enc}$; the current output tag $y_{i+1}$ is decoded by using a softmax loss function and is further fed as an input to the next time step. Finally, we obtain a tag sequence over all time steps.

\subsubsection{Pointer Networks}

Pointer networks apply RNNs to learn the conditional probability of an output sequence with elements that are discrete tokens corresponding to the positions in an input sequence~\cite{vinyals2015pointer,DBLPLiSJ18}. It represents variable length dictionaries by using a softmax probability distribution as a ``pointer''. Zhai \etal~\cite{zhai2017neural} first applied pointer networks to produce sequence tags.
Illustrated in Figure~\ref{sfig:pointer}, pointer networks first identify a chunk (or a segment), and then label it.
This operation is repeated until all the words in input sequence are processed.
In Figure~\ref{sfig:pointer}, given the start token ``<s>'', the segment ``Michael Jeffery Jordan'' is first identified and then labeled as ``PERSON''. The segmentation and labeling can be done by two separate neural networks in pointer networks.
Next,  ``Michael Jeffery Jordan''  is taken as input and fed into pointer networks. As a result, the segment ``was'' is identified and labeled as ``O''.

\subsection{Summary of DL-based NER}
\label{ssec:summaryDLNER}

\paratitle{Architecture Summary.} 
Table~\ref{tab:tabsurvey} summarizes recent works on neural NER by their architecture choices. 
BiLSTM-CRF is the most common architecture for NER using deep learning. 
The method \cite{DBLP:journals/corr/abs-1903-07785} which pre-trains a bi-directional Transformer model in a cloze-style manner, achieves the state-of-the-art performance ($93.5\%$) on CoNLL03.  
The work with BERT and dice loss \cite{DBLP:journals/corr/abs-1911-02855} achieves the state-of-the-art performance ($92.07\%$) on OntoNotes5.0.

The success of a NER system heavily relies on its input representation.
Integrating or fine-tuning pre-trained language model embeddings is becoming a new paradigm for neural NER. 
When leveraging these language model embeddings, there are significant performance improvements \cite{akbik2018contextual,peters2018deep,DBLP:conf/acl/LiuYL19,DBLP:conf/acl/XiaZYLDWFMY19,DBLP:journals/corr/abs-1910-11476,DBLP:journals/corr/abs-1911-02855,DBLP:journals/corr/abs-1911-02257,DBLP:conf/acl/LiuMZXCZ19,DBLP:journals/corr/abs-1903-07785,jiang2019improved}.
The last column in Table~\ref{tab:tabsurvey} lists the reported performance in F-score on a few benchmark datasets. While high F-scores have been reported on formal documents (\eg CoNLL03 and OntoNotes5.0), NER on noisy data (\eg W-NUT17) remains challenging. 

\paratitle{Architecture Comparison.} 
We discuss pros and cons from three perspectives: input, encoder, and decoder. 
First, no consensus has been reached about whether external knowledge  should be or how to integrate into DL-based NER models. 
Some studies \cite{zhuo2016segment,DBLP:conf/acl/LiuYL19,DBLP:conf/acl/XiaZYLDWFMY19,DBLP:journals/corr/abs-1909-10148} shows that NER performance can be boosted with external knowledge. 
However, the disadvantages are also apparent: 1) acquiring external knowledge is labor-intensive (\eg gazetteers) or computationally expensive (\eg dependency); 2) integrating external knowledge adversely affects end-to-end learning and hurts the generality of DL-based systems.

Second, Transformer encoder is more effective than LSTM when Transformer is pre-trained on huge corpora. 
Transformers fail on NER task if they are not pre-trained and when the training data is limited \cite{DBLP:conf/naacl/GuoQLSXZ19,yan2019tener}.
On the other hand, Transformer encoder is faster than recursive layers when the length of the sequence $n$ is smaller than the dimensionality of the representation $d$ (complexities: self-attention $\mathcal{O} (n^2 \cdot d)$ and recurrent $\mathcal{O} (n\cdot d^2)$) \cite{vaswani2017attention}.

Third, a major disadvantage of RNN and Pointer Network decoders lies in greedily decoding, which means that the input of current step needs the output of previous step.  
This mechanism may have a significant impact on the speed and  is an obstacle to parallelization. 
CRF is the most common choice for tag decoder.
CRF is powerful to capture label transition dependencies when adopting non-language-model (\ie non-contextualized) embeddings such as Word2vec and GloVe. 
However,  CRF could be computationally expensive when the number of entity types is large.
More importantly, CRF does not always lead to better performance compared with softmax classification when adopting contextualized language model embeddings such as BERT and ELMo \cite{DBLP:journals/corr/abs-1908-08676,DBLP:journals/corr/abs-1910-11476}.    

For end users, what architecture to choose is data and domain task dependent.  
If data is abundant, training models with RNNs from scratch and fine-tuning contextualized language models could be considered. 
If data is scarce, adopting transfer strategies might be a better choice. 
For newswires domain, there are many pre-trained off-the-shelf models available.
For specific domains (\eg medical and social media), fine-tuning general-purpose contextualized language models with domain-specific data is often an effective way.

\paratitle{NER for Different Languages.} In this survey, we mainly focus on NER in English and in general domain.
Apart from English language, there are many studies on other languages or cross-lingual settings. 
Wu \etal~\cite{wu2015named} and Wang \etal~\cite{wang2018incorporating} investigated NER in Chinese clinical text. Zhang and Yang~\cite{zhang2018chinese} proposed a lattice-structured LSTM model for Chinese NER, which encodes a sequence of input characters as well as all potential words that match a lexicon.  
Other than Chinese,  many studies are conducted on other languages. 
Examples  include Mongolian~\cite{wang2016mongolian}, Czech~\cite{strakova2016neural},  Arabic~\cite{gridach2016character}, Urdu~\cite{malik2017urdu}, Vietnamese~\cite{pham2017end}, Indonesian~\cite{kurniawan2018empirical}, and Japanese~\cite{yano2018neural}.
Each language has its own characteristics for understanding the fundamentals of NER task on that language.
There are also a number of studies~\cite{bharadwaj2016phonologically,yang2016multi,xie2018neural,DBLP:conf/acl/StoyanovJLY18} aiming to solve the NER problem in a cross-lingual setting by transferring knowledge from a source language  to a  target language with few or no labels.

\section{Applied Deep Learning for NER}
\label{sec:appliedDL}
Sections~\ref{ssec:input}, \ref{ssec:contextEncoder}, and \ref{ssec:tagDecoder} outline typical network architectures for NER. 
In this section, we survey recent applied deep learning techniques that are being explored for NER. 

\subsection{Deep Multi-task Learning for NER}

Multi-task learning~\cite{caruana1997multitask} is an approach that learns a group of related tasks together.
By considering the relation between different tasks, multi-task learning algorithms are expected to achieve better results than the ones that learn each task individually.

Collobert \etal~\cite{collobert2011natural} trained a window/sentence approach network  to jointly perform POS, Chunk, NER, and SRL tasks.
This multi-task mechanism lets the training algorithm to discover internal representations that are useful for all the tasks of interest.
Yang \etal~\cite{yang2016multi} proposed a multi-task joint model, to learn language-specific regularities, jointly trained for POS, Chunk, and NER tasks.
Rei~\cite{rei2017semi} found that by including an unsupervised language modeling objective in the training process, the sequence labeling model achieves consistent performance improvement.
Lin \etal \cite{DBLP:conf/acl/StoyanovJLY18} proposed a multi-lingual multi-task architecture for low-resource settings, which can effectively transfer different types of knowledge to improve the main model.

Other than considering NER together with other sequence labeling tasks, multi-task learning framework can be applied for joint extraction of entities and relations~\cite{zheng2017joint,zhou2017joint}, or to model NER as two related subtasks: entity segmentation and entity category prediction~\cite{peng2016multi,aguilar2017multi}.
In biomedical domain, because of the differences in different datasets, NER on each dataset is considered as a task in a multi-task setting~\cite{crichton2017neural,wang2018cross}. A main assumption here is that the different datasets share the same character- and word-level information. Then multi-task learning is applied to make more efficient use of the data and to encourage the models to learn
more generalized representations.

\subsection{Deep Transfer Learning for NER}
\label{ss:deeptransfer}

Transfer learning aims to perform a machine learning task on a target domain by taking advantage of  knowledge learned from a source domain~\cite{pan2010survey}.
In NLP, transfer learning is also known as domain adaptation. On NER tasks, the traditional approach is through  bootstrapping algorithms~\cite{jiang2007instance,wu2009domain,DBLP:journals/corr/abs-1908-08983}. Recently, a few approaches~\cite{pan2013transfer,lee2017transfer,lin2018neural,DBLP:journals/corr/abs-1908-09659,DBLP:conf/acl/JiaXZ19,DBLP:journals/corr/abs-1909-11535} have been proposed  for low-resource and across-domain NER using  deep neural networks.

Pan \etal~\cite{pan2013transfer} proposed a transfer joint embedding (TJE) approach for cross-domain NER. TJE employs label embedding techniques to transform  multi-class classification to regression in a low-dimensional latent space.  Qu \etal~\cite{qu2016named} observed that related named entity types often share lexical and context features.
Their approach learns the correlation between source and target named entity types using a two-layer neural network. Their approach is applicable to the setting that the source domain has similar (but not identical) named entity types with the target domain.  Peng and Dredze~\cite{peng2016multi} explored transfer learning in a multi-task learning setting, where they considered two domains: news and social media,  for two tasks: word segmentation and NER.

In the setting of transfer learning, different neural models commonly share different parts of model parameters between source task and target task.
Yang \etal~\cite{yang2017transfer} first investigated the transferability of different layers of representations.
Then they presented three different parameter-sharing architectures for cross-domain, cross-lingual, and cross-application scenarios.
If two tasks have mappable label sets, there is a shared CRF layer, otherwise, each task learns a separate CRF layer.
Experimental results show significant improvements on various datasets under low-resource conditions (\ie  fewer available annotations).
Pius and Mark~\cite{von2017transfer} extended Yang's approach to allow joint training on informal corpus (\eg WNUT 2017), and to incorporate sentence level feature representation.
Their approach achieved the 2nd place at the WNUT 2017 shared task for NER, obtaining an F1-score of 40.78\%.	
Zhao \etal~\cite{zhao2018improve} proposed a multi-task model with domain adaption, where the fully connection ayer are adapted to different datasets, and the CRF features are computed separately.
A major merit of Zhao's model is that the instances with different distribution and misaligned annotation guideline are filtered out in data selection procedure.
Different from these parameter-sharing architectures, Lee \etal~\cite{lee2017transfer} applied transfer learning in NER by training a model on source task and using the trained model on target task for fine-tuning.
Recently, Lin and Lu \cite{lin2018neural} also proposed a fine-tuning approach for NER by introducing three neural adaptation layers: word adaptation layer, sentence adaptation layer, and output adaptation layer.
Beryozkin \etal \cite{DBLP:conf/acl/BeryozkinDGHS19} proposed a tag-hierarchy model for  heterogeneous tag-sets NER setting, where the hierarchy is  used during inference to map fine-grained tags onto a target tag-set.  
 In addition, some studies~\cite{wang2018cross,giorgi2018transfer,DBLP:conf/naacl/WangQCSZZGGCY18} explored transfer learning in biomedical NER to reduce the amount of required labeled data.

\subsection{Deep Active Learning for NER}

The key idea behind active learning is that a machine learning algorithm can perform better with substantially less data from training, if it is allowed to choose the data from which it learns~\cite{settles2012active}. Deep learning typically requires a large amount of training data which is costly to obtain.
Thus, combining deep learning with active learning is expected to reduce data annotation effort.

Training with active learning proceeds in multiple rounds. However, traditional active learning schemes are expensive for deep learning because  after each round they require complete retraining of the classifier with newly annotated samples.
Because retraining from scratch is not practical for deep learning, Shen \etal~\cite{shen2017deep} proposed to carry out incremental training for NER with each batch of new labels. They mix newly annotated samples with the existing ones, and update neural network weights for a small number of epochs, before querying for labels in a new round.
Specifically, at the beginning of each round, the active learning algorithm chooses sentences to be annotated, to the predefined budget.
The model parameters are updated by training on the augmented dataset, after receiving chose annotations.
The active learning algorithm adopts uncertainty sampling strategy~\cite{lewis1994sequential}  in selecting sentences to be annotated.
Experimental results show that active learning algorithms achieve 99\% performance of the best deep learning model trained on full data using only 24.9\% of the training data on the English dataset and 30.1\% on Chinese dataset. Moreover,
12.0\% and 16.9\% of training data were enough for deep active learning model to outperform shallow models learned on full training data~\cite{pradhan2013towards}.

\subsection{Deep Reinforcement Learning for NER}

Reinforcement learning (RL) is a branch of machine learning inspired by behaviorist psychology, which is concerned with how software agents take actions in an environment so as to maximize some cumulative rewards~\cite{kaelbling1996reinforcement,sutton1998introduction}.
The idea is that an agent will learn from the environment by interacting with it and receiving rewards for performing actions.
Specifically, the RL problem can be formulated as follows~\cite{hoi2018online}:
the environment is modeled as a stochastic finite state machine with inputs (actions from agent) and outputs (observations and rewards to the agent). It consists of three key components: (i) state transition function, (ii) observation (\ie output) function, and (iii) reward function.
The agent is also modeled as a stochastic finite state machine with inputs (observations/rewards from the environment) and outputs (actions to the environment). It consists of two components: (i) state transition function, and (ii) policy/output function.
The ultimate goal of an agent is to learn a good state-update function and policy by attempting to maximize the cumulative rewards.

Narasimhan \etal~\cite{narasimhan2016improving}  modeled the task of information extraction as a Markov decision process (MDP), which dynamically incorporates entity predictions and provides flexibility to choose the next search query from a set of automatically generated alternatives. The process is comprised of issuing search queries, extraction from new sources, and reconciliation of extracted values, and the process repeats until sufficient evidence is obtained.
In order to learn a good policy for an agent, they utilize a deep Q-network~\cite{mnih2015human} as a function approximator, in which the state-action value function (\ie Q-function) is approximated by using a deep neural network.
Recently, Yang \etal \cite{DBLP:conf/coling/YangCLHZ18} utilized the data generated by distant supervision to perform new type named entity recognition in new domains. 
The instance selector is based on reinforcement learning and obtains the feedback reward from the NE tagger, aiming at choosing positive sentences to reduce the effect of noisy annotation.

\subsection{Deep Adversarial Learning for NER}

Adversarial learning~\cite{lowd2005adversarial} is the process of explicitly training a model on adversarial examples. The purpose is to make the model more robust to attack or to reduce its test error on clean inputs.
Adversarial networks learn to generate from a training distribution through a 2-player game: one network generates candidates (generative network) and the other evaluates them (discriminative network).
Typically, the generative network learns to map from a latent space to a particular data distribution of interest, while the discriminative network discriminates between candidates generated by the generator and instances from the real-world data distribution \cite{goodfellow2014generative}.

For NER, adversarial examples are often produced in two ways. 
Some studies \cite{DBLP:conf/naacl/HuangJM19,DBLP:conf/ijcai/0034YS19,DBLP:conf/emnlp/Cao000L18} considered the instances in a source domain as adversarial examples for a target domain, and vice versa. 
For example, 
Li \etal \cite{DBLP:conf/ijcai/0034YS19} and  Cao \etal \cite{DBLP:conf/emnlp/Cao000L18} both incorporated adversarial examples from other domains to encourage domain-invariant features for cross-domain NER. 
Another option is to prepare an adversarial sample by adding an original sample with a perturbation. 
For example, dual adversarial transfer network (DATNet), proposed in~\cite{anonymous2019datnet},  aims to deal with the problem of low-resource NER.
An adversarial sample is produced by adding original sample with a perturbation bounded by a small norm $\epsilon$ to maximize the loss function as follows: ${\eta _x} = \arg \mathop {\max }\limits_{\eta :{{\left\| \eta  \right\|}_2} \le \epsilon } l(\Theta ;x + \eta )$, 
where $\Theta$ is the current model parameters set, $\epsilon$ can be determined on validation set.
An adversarial example is constructed by $x_{adv}= x +\eta _x$.
The classifier is trained on the mixture of original and adversarial examples to improve generalization.

\subsection{Neural Attention for NER}

The attention mechanism  is loosely based on the visual attention mechanism found in human \cite{britz2016attention}.
For example, people usually focus on a certain region of an image with ``high resolution'' while perceiving the surrounding region with ``low resolution''.
Neural attention mechanism allows neural networks have the ability to focus on a subset of its inputs.
By applying attention mechanism, a NER model could capture the most informative elements in the inputs. In particular, the Transformer architecture reviewed in Section~\ref{sssec:transformer} relies entirely on attention mechanism to draw global dependencies between input and output.

There are many other ways of applying attention mechanism in NER tasks. Rei \etal~\cite{rei2016attending} applied an attention mechanism to dynamically decide how much information to use from a character- or word-level component in an end-to-end NER model.
Zukov-Gregoric \etal~\cite{zukov2017neural} explored the self-attention mechanism in NER, where the
weights are dependent on a single sequence (rather than on the relation between two sequences).
Xu \etal~\cite{xu2018improving} proposed an attention-based neural NER architecture to leverage document-level global information.
In particular, the document-level information  is obtained from document represented by pre-trained bidirectional language model with neural attention. Zhang \etal~\cite{zhang2018adaptive} used an adaptive co-attention network for NER in tweets.
This adaptive co-attention network is a multi-modal model using co-attention process. Co-attention includes visual attention and textual attention to capture the semantic interaction between different modalities.

\section{Challenges and Future Directions}
\label{sec:insights}

Discussed in Section~\ref{ssec:summaryDLNER}, the choices tag decoders do not vary as much as the choices of input representations and context encoders. From Google Word2vec to the more recent BERT model, DL-based NER benefits significantly from the advances made in pre-trained embeddings in modeling languages.  Without the need of complicated feature-engineering, we now have the opportunity to re-look the NER task for its challenges and potential future directions.

\subsection{Challenges}
\label{ssec:challenges}

\paratitle{Data Annotation.} Supervised NER systems, including DL-based NER, require big annotated data in training. However, data annotation remains time consuming and expensive. It is a big challenge for many resource-poor languages and specific domains as domain experts are needed to perform annotation tasks.

Quality and consistency of the annotation are both major concerns because of the language ambiguity. For instance, a same named entity may be annotated with different types. As an example, ``\textit{Baltimore}'' in the sentence ``\textit{Baltimore defeated the Yankees}'', is labeled as Location in MUC-7 and Organization in CoNLL03. Both ``\textit{Empire State}'' and ``\textit{Empire State Building}'', is labeled as Location in CoNLL03 and ACE datasets, causing confusion in entity boundaries. Because of the inconsistency in data annotation, model trained on one dataset may not work well on another even if the documents in the two datasets are from the same domain.

To make data annotation even more complicated, Katiyar and Cardie~\cite{katiyar2018nested} reported that nested entities are fairly common: 17\% of the entities in the GENIA corpus are embedded within another entity; in the ACE corpora, 30\% of sentences contain nested entities. There is a need to develop common annotation schemes to be applicable to both nested entities and fine-grained entities, where one named entity may be assigned multiple types.

\paratitle{Informal Text and Unseen Entities.} Listed in Table~\ref{tab:tabsurvey}, decent results are reported on datasets with formal documents (\eg news articles). However, on user-generated text \eg WUT-17 dataset, the best F-scores are slightly above 40\%. NER on informal text (\eg tweets, comments, user forums) is more challenging than on formal text due to the shortness and noisiness. Many user-generated texts are domain specific as well. In many application scenarios, a NER system has to deal with user-generated text such as customer support in e-commerce and banking.

Another interesting dimension to evaluate the robustness and effectiveness of NER system is its capability of  identifying unusual, previously-unseen entities in the context of emerging discussions. There exists a shared task\footnote{\url{https://noisy-text.github.io/2017/emerging-rare-entities.html}} for this direction of research on WUT-17 dataset~\cite{derczynski2017results}.

\subsection{Future Directions}

With the advances in modeling languages and demand in real-world applications, we expect NER to receive more attention from researchers.  On the other hand, NER is in general considered as a pre-processing component to downstream applications. That means a particular NER task is defined by the requirement of downstream application, \eg the types of named entities and whether there is a need to detect nested entities \cite{DBLP:conf/acl/FisherV19}. Based on the studies in this survey, we list the following directions for further exploration in NER research.

\paratitle{Fine-grained NER and Boundary Detection.} While many existing studies \cite{ma2016end,lample2016neural,ghaddar2018robust} focused on  coarse-grained NER in general domain, we expect more research on fine-grained NER in domain-specific areas to support various real word applications \cite{DBLP:conf/wcre/YeXFALK16}. The challenges in fine-grained NER are the significant increase in NE types and the complication introduced by allowing a named entity to have multiple NE types. This calls for a re-visit of the common NER approaches where the entity boundaries and the types are detected simultaneously \eg by using B- I- E- S-(entity type) and O as the decoding tags.
It is worth considering to define \textit{named entity boundary detection} as a dedicated task to detect NE boundaries while ignoring the NE types. The decoupling of boundary detection and NE type classification enables common and robust solutions for boundary detection that can be shared across different domains, and dedicated domain-specific approaches for NE type classification. Correct entity boundaries also effectively alleviate error propagation in entity linking to knowledge bases.
There has been some studies \cite{partalas2016learning,zhai2017neural} which consider entity boundary detection as an intermediate step (\ie a subtask) in NER.
To the best of our knowledge, no existing work separately focuses on entity boundary detection to provide a robust recognizer.
We expect a breakout in this research direction in the future.

\paratitle{Joint NER and Entity Linking.}
Entity linking (EL) \cite{shen2018shine}, also referred to as named entity normalization or disambiguation, aims at assigning a unique identity to entities mentioned in text with reference to a knowledge base, \eg Wikipedia in general domain and the Unified Medical Language System (UMLS) in biomedical domain.
Most existing works individually solve NER and EL as two separate tasks in a pipeline setting. 
We consider that the semantics carried by the successfully linked entities (\eg through the related entities in the knowledge base) are significantly enriched \cite{ji2016joint,phan2018pair}. That is, linked entities contributes to the successful detection of entity boundaries and correct classification of entity types. It is worth exploring approaches for jointly performing NER and EL, or even entity boundary detection, entity type classification, and entity linking, so that each subtask benefits from the partial output by other subtasks, and alleviate error propagations that are unavoidable in pipeline settings.

\paratitle{DL-based NER on Informal Text with Auxiliary Resource.}
As discussed in Section~\ref{ssec:challenges}, performance of DL-based NER on informal text or user-generated content remains low. This calls for more research in this area. In particular, we note that the performance of NER benefits significantly from the availability of  auxiliary resources \cite{li2017extracting,han2018linking,phan2019collective}, \eg a dictionary of location names in user language. While Table~\ref{tab:tabsurvey} does not provide strong evidence of involving gazetteer as additional features leads to performance increase to NER in general domain, we consider auxiliary resources are often necessary to better understand user-generated content. The question is how to obtain matching auxiliary resources for a NER task on user-generated content or domain-specific text, and how to effectively incorporate the auxiliary resources in DL-based NER.

\paratitle{Scalability of DL-based NER}.
Making neural NER models more scalable is still a challenge. 
Moreover, there is still a need for solutions on optimizing exponential growth of parameters when the size of data grows \cite{batmaz2018review}.
Some DL-based NER models have achieved good performance with the cost of massive computing power.
For example, the ELMo representation represents each word with a $3\times1024$-dimensional vector, and the model was trained for 5 weeks on 32 GPUs~\cite{akbik2018contextual}.
Google BERT representations were trained on 64 cloud TPUs.
However,  end users are not able to fine-tune these models if they have no access to powerful computing resources.
Developing approaches to balancing model complexity and scalability will be a promising direction.
On the other hand, model compression and pruning techniques are also options to reduce the space and computation time required for model learning.

\paratitle{Deep Transfer Learning for NER}.
Many entity-focused applications resort to off-the-shelf NER systems to recognize named entities.
However, model trained on one dataset may not work well on other texts due to the differences in characteristics of languages as well as the differences in annotations.
Although there are some studies of applying deep transfer learning to NER  (see Section~\ref{ss:deeptransfer}), this problem has not been fully explored.
More future efforts should be dedicated on how to effectively transfer knowledge from one domain to another by exploring the following research problems:
(a) developing a robust recognizer, which is able to work well across different domains;
(b) exploring zero-shot, one-shot and few-shot learning in NER tasks;
(c) providing solutions to address domain mismatch, and label mismatch in cross-domain settings.

\paratitle{An Easy-to-use Toolkit for DL-based NER}.
Recently, R{\"o}der \etal \cite{roder2017gerbil} developed GERBIL, which provides researchers, end users and developers with easy-to-use interfaces  for benchmarking entity annotation tools with the aim of ensuring repeatable and archiveable experiments.
However, it does not involve recent DL-based techniques.
Ott \cite{DBLP:conf/naacl/OttEBFGNGA19} presented FAIRSEQ, a fast, extensible toolkit for sequence modeling, especially for machine translation and text stigmatization.
Dernoncourt \etal \cite{2017neuroner} implemented a framework, named NeuroNER, which only relies on a variant of recurrent neural network.
In recent years, many deep learning frameworks (\eg TensorFlow, PyTorch, and Keras) have been designed to offer building blocks for designing, training and validating deep neural networks, through a high level programming interface.\footnote{\url{https://developer.nvidia.com/deep-learning-frameworks}}
In order to re-implement the architectures in Table~\ref{tab:tabsurvey}, developers may write codes from scratch with existing deep learning frameworks.
We envision that an easy-to-use NER toolkit can guide developers to complete it with some standardized modules: data-processing, input representation, context encoder, tag decoder, and effectiveness measure. We believe that experts and non-experts can both benefit from such toolkits.

\section{Conclusion}
\label{sec:conclusion}

This survey aims to review recent studies on deep learning-based NER solutions to help new researchers building a comprehensive understanding of this field. We include in this survey the background of the NER research, a brief of traditional approaches, current state-of-the-arts,  and challenges and future research directions.
First, we consolidate available NER resources, including tagged NER corpora and off-the-shelf NER systems, with focus on NER in general domain and NER in English.  We present these resources in a tabular form and provide links to them for easy access.
Second, we introduce  preliminaries such as definition of NER task, evaluation metrics,  traditional approaches to NER, and basic concepts in deep learning.
Third, we review the literature based on varying models of deep learning and map these studies according to a new taxonomy.
We further survey the most representative methods for recent applied deep learning techniques in new problem settings and applications.
Finally, we summarize the applications of NER and present readers with challenges in NER and future directions.
We hope that this survey can provide a good reference when designing DL-based NER models.

\bibliographystyle{IEEEtran}
\bibliography{ner}

\vfill

\end{document}